\documentclass[11pt]{article}

\usepackage[final]{acl}

\usepackage{times}
\usepackage{latexsym}

\usepackage[T1]{fontenc}

\usepackage[utf8]{inputenc}

\usepackage{microtype}
\usepackage{hyperref}
\usepackage{url}
\usepackage{booktabs}
\usepackage{xspace}
\usepackage{graphicx}
\usepackage{textcomp}
\usepackage{multicol}
\usepackage{multirow}
\usepackage{enumitem}
\usepackage{pifont}
\usepackage{amsmath}
\usepackage{amsfonts}
\usepackage{subfigure}
\usepackage{float}
\usepackage{tikz}
\usepackage{subcaption}
\usepackage{lineno}
\usepackage{algorithm}
\usepackage{algpseudocode}
\usepackage{algorithmicx}
\usepackage{xcolor}
\usepackage[table]{xcolor}
\definecolor{Gray2}{gray}{0.7}
\usepackage{inconsolata}
\usepackage{amssymb}
\usepackage{soul}
\definecolor{lightyellow}{rgb}{1, 1, 0.8}
\sethlcolor{lightyellow}
\usepackage{listings}

\newcommand{\cmark}{{\color{green!60!black}\ding{51}}}
\newcommand{\xmark}{{\color{red!75!black}\ding{55}}}

\usepackage{graphicx}

\definecolor{darkblue}{rgb}{0, 0, 0.5}
\hypersetup{colorlinks=true, citecolor=darkblue, linkcolor=darkblue, urlcolor=darkblue}

\def \name{\textsc{SPRInG}\xspace}
\definecolor{mycolor}{RGB}{33, 95, 154}

\title{SPRInG: Continual LLM Personalization via Selective Parametric Adaptation and Retrieval-Interpolated Generation}

\author{
  Seoyeon Kim \qquad
  Jaehyung Kim \\
  Yonsei University \\
  \texttt{\{yeon\_04, jaehyungk\}@yonsei.ac.kr}
}

\begin{document}
\maketitle
\nolinenumbers

\begin{abstract}
Personalizing Large Language Models typically relies on static retrieval or one-time adaptation, assuming user preferences remain invariant over time.
However, real-world interactions are dynamic, where user interests continuously evolve, posing a challenge for models to adapt to preference drift without catastrophic forgetting.
Standard continual learning approaches often struggle in this context, as they indiscriminately update on noisy interaction streams, failing to distinguish genuine preference shifts from transient contexts.
To address this, we introduce \name{}, a novel semi-parametric framework designed for effective continual personalization.
During training, \name{} employs drift-driven selective adaptation, which utilizes a likelihood-based scoring function to identify high-novelty interactions, selectively updating the user-specific adapter on drift signals while preserving hard-to-learn residuals in a replay buffer.
During inference, we apply strict relevance gating and fuse parametric knowledge with retrieved history via probability interpolation.
Experiments on the long-form personalized generation benchmark demonstrate that \name{} significantly outperforms existing baselines, validating its robustness for real-world continual personalization.\looseness=-1\footnote{The code is available at \url{https://github.com/ssseoyneee/EMNLP26_SPRInG}.}
\end{abstract}

\section{Introduction}

\begin{figure}[ht]
\centering
\includegraphics[width=\columnwidth]{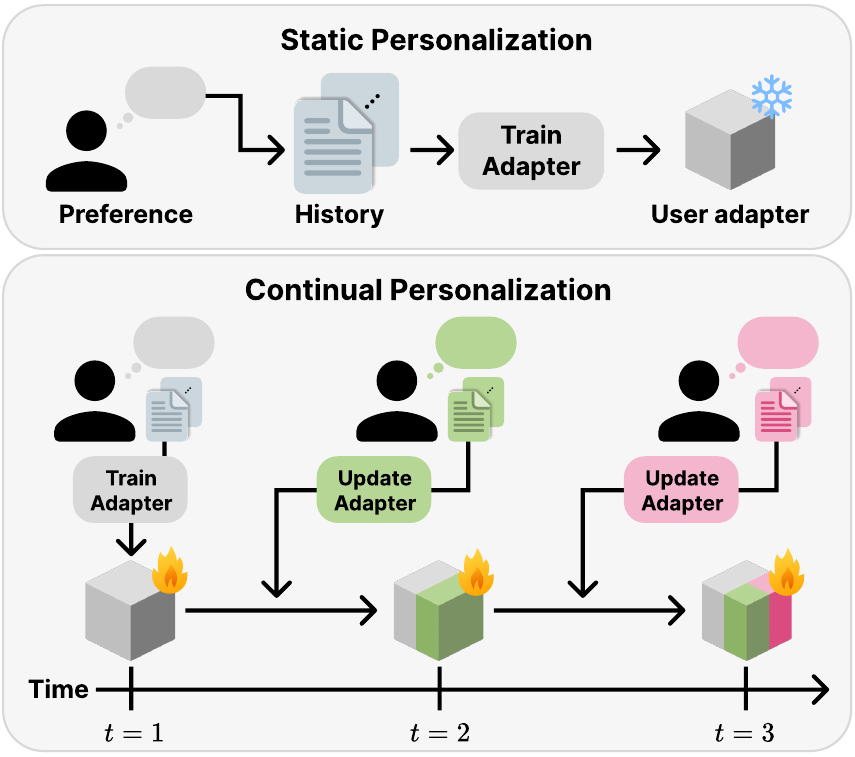}
\vspace{-0.2in}
\caption{
\textbf{ 
Comparison between static and continual personalization.} 
\textbf{(Top)} Static personalization assumes time-invariant preferences, using a frozen adapter trained on a fixed history snapshot.
\textbf{(Bottom)} Continual personalization models users as dynamic entities. The framework iteratively updates the adapter at sequential time steps ($t=1, 2, 3, \cdots $) to capture preference drifts (shifting colors) while maintaining historical context.
}\label{fig:intro_concept}
\vspace{-0.2in}
\end{figure}

\begin{figure*}[!t]
\centering
\includegraphics[width=1.0\textwidth]{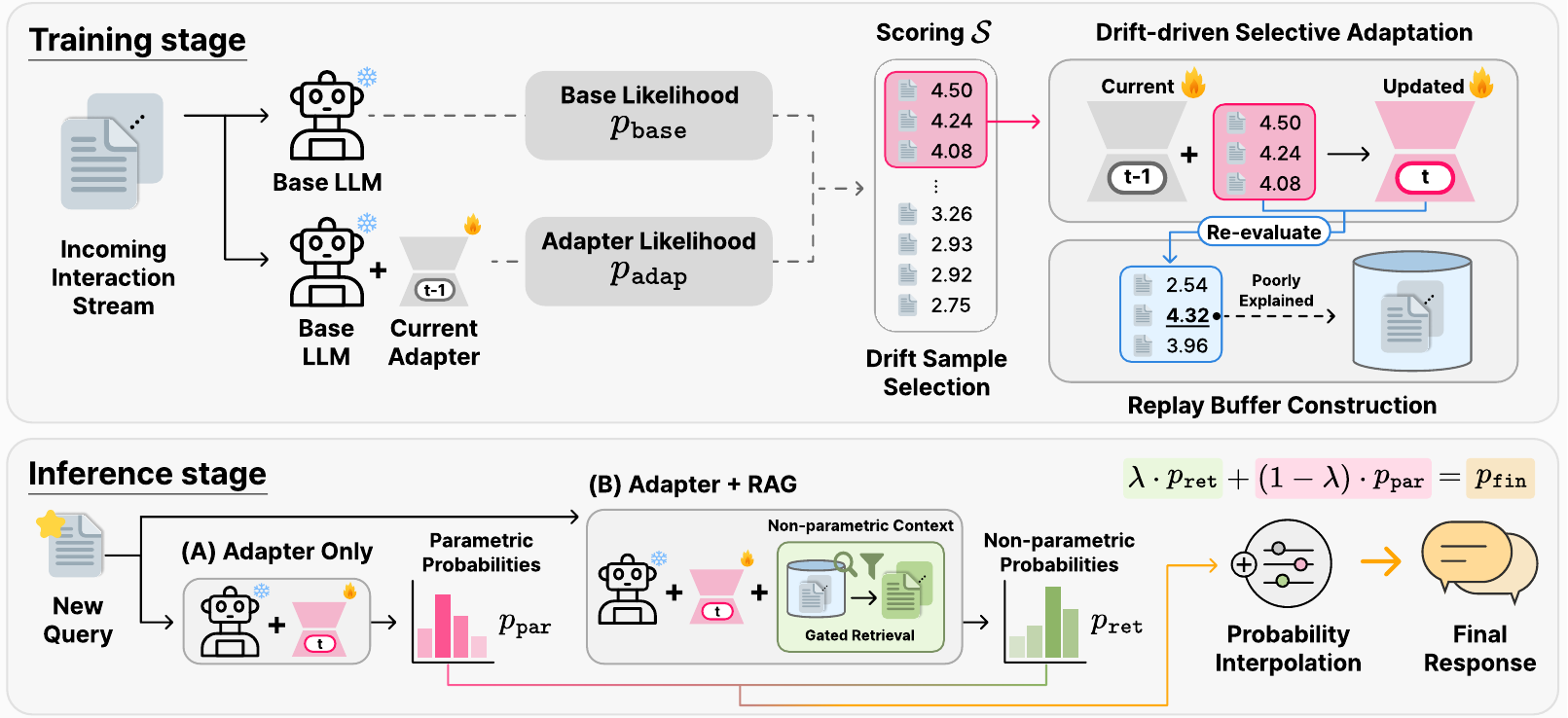}
\caption{
\textbf{An overview of \name{}.}
\textbf{Training stage:} incoming interactions are scored by comparing likelihoods between the base and adapted models. High-scoring \textit{drift samples} trigger a parametric update, while samples that remain poorly internalized are stored in the replay buffer through re-evaluation with respect to the updated model.
\textbf{Inference stage:} for a new query, the model computes parametric probabilities ($p_{\texttt{par}}$) from the adapter and non-parametric probabilities ($p_{\texttt{ret}}$) via \textit{gated retrieval}. The final output is generated through \textit{probability interpolation}, dynamically weighting the two distributions to balance internal knowledge with retrieved context.
}
\vspace{-0.1in}
\label{fig:method}
\end{figure*}

While Large Language Models (LLMs) demonstrate remarkable performance across diverse domains \citep{jaech2024openai, comanici2025gemini, yang2025qwen3}, the need for \textit{LLM personalization}, which adapts outputs to individual or group preferences, is rapidly increasing \citep{zhang2024personalization}. 
Existing LLM personalization methods adopt either non-parametric strategies that leverage retrieved interaction histories or user profiles during prompting \citep{santurkar2023whose, hwang2023aligning}, or parametric strategies that encode user preferences through parameter adaptation (\textit{e.g.,} training user-specific LoRA adapters \citep{tan2024democratizing, bu2025personalized}). 
However, both paradigms assume a \textit{time-invariant view} of the user, in that a fixed history is repeatedly used for retrieval and parametric models are generally frozen after initial training.
Yet, in reality, user preferences and traits constantly evolve over time, causing existing methods to fail to account for such \textit{preference drifts} (Figure~\ref{fig:intro_concept}).
To be more specific, non-parametric methods struggle with expanding histories due to weak recency signals and rising costs of storage and retrieval, whereas parametric approaches require costly updates and a risk of catastrophic forgetting.
This limitation necessitates the development of \textit{continual LLM personalization} that accounts for users' changing states over time.

\textit{Continual learning} (CL) \citep{ke2022continual, shi2024continual}, which aims to update models sequentially with streaming data, offers a promising paradigm for handling evolving user preferences. 
Prior CL research has primarily concentrated on mitigating catastrophic forgetting in task-incremental settings, typically relying on clear task boundaries or abrupt distribution shifts to determine \textit{when} and \textit{how} to update \citep{du2024unlocking, kirkpatrick2017overcoming}.
However, continual LLM personalization poses distinct challenges; user interaction streams are inherently continuous and noisy, where preferences drift gradually rather than exhibiting discrete shifts.
The central challenge, therefore, lies in \textit{distinguishing genuine preference evolution from transient noise}. 
In this context, standard CL approaches could be suboptimal due to this discrepancy; for example, transient noise could be considered informative from a CL perspective (\textit{e.g.}, due to high loss or uncertainty).
This necessitates a framework for selective drift detection, ensuring the model adapts to meaningful new traits while robustly filtering out noise.

To bridge this gap, we propose \textbf{\name{}} (\textbf{S}elective \textbf{P}arametric adaptation and \textbf{R}etrieval-\textbf{In}terpolated \textbf{G}eneration), a novel semi-parametric framework designed to continuously adapt to evolving user preferences while preserving important historical contexts.
Our key insight is to address continual personalization across both training and inference stages, jointly utilizing parametric and non-parametric approaches for distinct yet complementary roles.
Specifically, during training, we employ \textit{Drift-Driven Selective Adaptation}, using a new scoring function to measure sample novelty relative to the current adapter and fine-tuning the model exclusively on interactions that indicate genuine preference drift.
Simultaneously, we construct a residual replay buffer by identifying hard samples that remain challenging even after parametric adjustments, preserving sparse but critical historical nuances that the adapter fails to internalize.
At inference, we apply \textit{Selective Retrieval-Interpolated Generation}; instead of naive retrieval-augmented generation (RAG), we introduce a strict gating mechanism to filter out irrelevant context and dynamically fuse parametric adapter knowledge with non-parametric replay-buffer evidence through probability-level interpolation.
This hybrid design ensures that \name{} remains both stable over time and responsive to the user's evolving preferences.

We demonstrate the effectiveness of \name{} by evaluating personalized long-form text generation tasks (Abstract Generation and Review Writing) from the LongLaMP benchmark \citep{kumar2024longlamp}. 
Empirical results show that \name{} consistently outperforms comprehensive baselines, surpassing the strongest CL method, CAMA \citep{kim2024online}, by a substantial margin of 15.85\% in ROUGE-L and standard RAG by 18.12\% in Review Writing.
Furthermore, our analyses validate the effectiveness of our design choices, showing an 8.0\% gain from drift-driven selective adaptation over updating on all interactions and an improvement of up to 23.2\% from retrieval-interpolated generation over the adapter-only setting.
Overall, these results validate \name{} as a robust framework for continual personalization, effectively reconciling the challenges of adapting to evolving preferences while preserving historical stability.
\section{Problem Formulation}

We formulate the personalized text generation task in a CL setup. 
Let a user $u$ interact with an LLM sequentially over time, producing an interaction history $\mathcal{D}_{u}=\{(q_i, a_i)\}$, where $q_i$ and $a_i$ denote the query and response for the $i$-th interaction, respectively. 
We employ a frozen base LLM $\mathcal{M}_\theta$ equipped with a lightweight user-specific adapter $\phi_u$ to capture user preferences.
We assume that accumulated interactions arrive sequentially over discrete time steps (periods) $t$, with the initial history given at $t=0$.
At each subsequent period $t=1, \ldots,T$, a new batch of interactions $\mathcal{D}_u^{(t)}$ becomes available. 
The objective is then to update the user adapter from $\phi_u^{(t-1)}$ to $\phi_u^{(t)}$ to reflect the evolving user preferences encoded in $\mathcal{D}_u^{(t)}$, while maintaining a non-parametric retrieval buffer $\mathcal{B}_u^{(t)}$ to preserve important past information.
We denote the user-adapted model at period $t$ as $\mathcal{M}_u^{(t)}$, which is the frozen base model $\mathcal{M}_\theta$ equipped with the current user adapter $\phi_u^{(t)}$.
With the user-adapted model, the generation of response $a$ for a given query $q$ is modeled by conditional probability $P(a|q;\mathcal{M}_u^{(t)})$.

Standard CL approaches generally update the adapter $\phi_u^{(t)}$ by minimizing the negative log-likelihood over the \textit{entire} incoming stream $\mathcal{D}_u^{(t)}$. 
However, such an indiscriminate update strategy is inappropriate for capturing \textit{subtle preference drift}.
Since user interactions are often noisy or context-dependent, treating all samples equally could obscure meaningful signals and cause the model to overfit to temporary contexts rather than genuine preference shifts. 
To address this, we propose a method that \textit{selectively} utilizes interactions indicative of true preference drift, thereby achieving effective personalization while filtering out noise. 
\section{Selective Parametric Adaptation and Retrieval-Interpolated Generation}

In this section, we propose \name{} (\textbf{S}elective \textbf{P}arametric adaptation and \textbf{R}etrieval-\textbf{In}terpolated \textbf{G}eneration), which conducts a two-phase process for continual LLM personalization.
First, drift-driven selective adaptation (Sec. \ref{sec:period_wise_cl}) \textit{selectively} updates the adapter by prioritizing interactions with high drift scores, capturing genuine preference shifts while retaining critical interactions in a replay buffer.
Second, selective retrieval-interpolated generation (Sec. \ref{sec:hybrid_inference}) dynamically fuses this parametric knowledge with historical evidence during inference.
This separation ensures that the adapter focuses on learning persistent preference patterns, leaving complex nuances to be handled by the retrieval memory.
See Figure~\ref{fig:method} for an overview.

\subsection{Training: Drift-driven selective adaptation and residual replay}
\label{sec:period_wise_cl}
The training phase of \name{} is designed to efficiently capture user preferences from the incoming interaction stream through a three-step pipeline. 
First, we perform \textit{drift selection} by comparing likelihoods between the base LLM and the same model equipped with the current adapter to isolate high-novelty interactions. 
Next, we execute a \textit{parametric update}, fine-tuning the user adapter on these identified samples to internalize genuine preference shifts. 
Finally, we construct a \textit{residual replay buffer} by re-evaluating the samples post-update; interactions that remain challenging are stored in the buffer to mitigate forgetting for the inference phase.
The overall procedure is summarized in Algorithm~\ref{alg:training}.

\paragraph{Drift sample selection.} 
To identify interactions to use for an adapter update, we define a scoring function that quantifies sample novelty with respect to the current adapter, thereby distinguishing genuine preference drift from uninformative noise. 
Let $p_{\texttt{base}} = P(a \mid q;\mathcal{M}_\theta)$ denote the likelihood of the response $a$ under the base model, and $p_{\texttt{adap}}^{(t-1)} = P(a \mid q; \mathcal{M}_u^{(t-1)})$ denote the likelihood under model adapted with the current adapter.
For both models, we compute $\log p$ as the mean token log-likelihood under the same maximum sequence length, preventing a systematic preference for shorter responses. 
Then, we define \textit{drift score} $\mathcal{S}(q, a)$ as follows:
\begin{equation}
    \label{eq:drift_score}
    \mathcal{S}(q, a) = (\underbrace{- \log \frac{p_{\texttt{adap}}^{(t-1)}}{p_{\texttt{base}}}}_{\text{Novelty Term}}) + \underbrace{\alpha \log p_{\texttt{base}}}_{\text{Quality Term}}
\end{equation}
where $\alpha$ is a balancing hyperparameter. 
We prioritize samples that maximize this score, balancing two critical objectives.
\begin{itemize}[leftmargin=3.5mm, topsep=3pt, itemsep=3pt, parsep=0pt]
    \item[$\circ$] The \textit{Novelty Term} assigns higher priority to samples with unfamiliar preference patterns.
    Thus, if the adapter already outperforms the base model ($p_{\texttt{adap}} \gg p_{\texttt{base}}$), the novelty score decreases as the preference is considered already internalized.
    \item[$\circ$] The \textit{Quality Term} filters for linguistic validity, using a high $p_{\texttt{base}}$ to identify well-formed and intelligible samples. 
    This term acts as a precautionary safeguard against \textit{pseudo-novelty}, preventing noisy outliers or gibberish from being erroneously selected simply because they are difficult for the adapter to predict.
\end{itemize}
Finally, we construct the drift set $\mathcal{D}_{\texttt{drift}}^{(t)}$ by retaining the top $p\%$ of the samples:
\begin{equation}
    \mathcal{D}_{\texttt{drift}}^{(t)} = \{ (q, a) \in \mathcal{D}_u^{(t)} \mid \mathcal{S}(q, a) \geq \tau_p \}
\end{equation}
where $\tau_p$ denotes the corresponding score cutoff within the current batch $\mathcal{D}_u^{(t)}$.

\paragraph{Parametric adapter update.} 
We employ Low-Rank Adaptation (LoRA) \citep{hu2022lora} to efficiently evolve the user model, which has been demonstrated in the context of LLM personalization \citep{tan2024democratizing}. 
For the initial period ($t=0$), we train $\phi_u^{(0)}$ on the observed portion of $\mathcal{D}_u^{(0)}$. 
For subsequent periods ($t > 0$), we initialize the current adapter with parameters from the previous period ($\phi_u^{(t)} \leftarrow \phi_u^{(t-1)}$) to inherit prior user preferences.
We then fine-tune $\phi_u^{(t)}$ \textit{exclusively} on the selected drift set $\mathcal{D}_{\texttt{drift}}^{(t)}$. 
By restricting training to high-novelty samples, we force the adapter to focus solely on acquiring new preference patterns. 
This targeted approach not only minimizes computational overhead but also prevents the model from overfitting to well-internalized stationary data.

\paragraph{Residual replay buffer construction.} 
Since parametric updates favor generalizable patterns, weight-encoded knowledge often misses sparse but critical historical nuances.
To mitigate this, we maintain a non-parametric replay buffer $\mathcal{B}_u$ as a complementary memory, preserving interactions that remain challenging for the updated adapter.

At period $t$, we first merge the newly identified drift samples with the existing buffer to form a candidate pool:
\begin{equation}
    \mathcal{C}_u^{(t)} = \mathcal{D}_{\texttt{drift}}^{(t)} \cup \mathcal{B}_u^{(t-1)}.
\end{equation}
We then \emph{re-evaluate} all samples in $\mathcal{C}_u^{(t)}$ using the \textit{updated} adapter $\phi_u^{(t)}$; we compute the post-update score
$\mathcal{S}'_t(q,a) := \mathcal{S}(q,a;\mathcal{M}_\theta,\phi_u^{(t)})$,
which quantifies the \textit{residual difficulty} after the parametric update.
This step is crucial because many drift samples that initially appear novel may become well-internalized after the adapter is updated; re-scoring allows us to filter out such redundant data and find samples that remain poorly predicted.

Finally, we update the replay buffer $\mathcal B_u^{(t)}$ under a fixed budget $N_{\max}$ using a \textit{Global Highest} retention policy. Specifically, when the candidate pool exceeds the budget, we rank all candidates by their post-update scores $\mathcal{S}_t'$ and retain those with the highest scores (Eq.~\ref{eq:buffer}).
We compare this policy with alternative retention strategies in Appendix~\ref{sec:appendix_buffer}.
\begin{equation}
\label{eq:buffer}
\small
\hspace{-20pt}
\mathcal{B}_u^{(t)} = 
\begin{cases} 
\mathcal{C}_u^{(t)},
& \text{if } |\mathcal{C}_u^{(t)}| \leq N_{\max} \\[2pt]
\operatorname{TopK}_{N_{\max}}
\left(\mathcal{C}_u^{(t)}; \mathcal{S}'_t\right),
& \text{otherwise}
\end{cases}
\end{equation}

\subsection{Inference: Gated replay-augmented generation and probability interpolation}
\label{sec:hybrid_inference}

In the inference stage, \name{} employs a \textit{Token-level Probability Interpolation} strategy, which dynamically balances the parametric knowledge of $\mathcal{M}_u^{(t)}$ with an explicit non-parametric context from the replay buffer $\mathcal{B}_u^{(t)}$ at each decoding step. 
This inference-time approach effectively addresses the limitation where parametric weights often fail to capture specific recent contexts or nuances that have not yet been fully internalized, whereas standard retrieval risks introducing irrelevant noise.
The overall procedure is presented in Algorithm~\ref{alg:inference}.

\paragraph{Relevance gating mechanism.}
Given a test query $q$, we first retrieve a candidate set $\mathcal R_q$ containing the top-$k$ relevant interactions from $\mathcal{B}_u^{(t)}$.
To prevent the detrimental effects of irrelevant retrieved contexts, we introduce a \textit{relevance gating mechanism} using a threshold $\sigma$.
Specifically, we define $\mathbf{c}_{\texttt{ret}}$ as the concatenation of gated interactions from $\mathcal {R}_q$:
\begin{equation}
\label{eq:gating}
    \mathbf{c}_{\texttt{ret}} = \textsc{Concat} \{ h \in \mathcal{R}_q \mid \textsc{Sim}(q, h) > \sigma \}
\end{equation}
where $\textsc{Sim}(\cdot)$ is a retrieval function.
This mechanism ensures that the model is augmented \textit{only when strong historical evidence exists}.
If no documents pass, $\mathbf{c}_{\texttt{ret}}$ remains empty, reverting to purely parametric generation.

\paragraph{Dual-path probability interpolation.} 
To dynamically balance internalized preferences with retrieved evidence during decoding steps, we employ a probability interpolation strategy between two different paths using the same user-adapted model $\mathcal{M}_u^{(t)}$.
\begin{itemize}[leftmargin=3.5mm, topsep=3pt, itemsep=3pt, parsep=0pt]
    \item[$\circ$] \textit{Parametric Path} takes only the query $q$ as input to $\mathcal{M}_u^{(t)}$, relying solely on encoded knowledge in adapter weights to guide generation.
    \item[$\circ$] \textit{Retrieval-augmented Path} inputs the query $q$ augmented with the retrieved context $\mathbf{c}_{\texttt{ret}}$, providing the model with explicit historical evidence.
\end{itemize}
Specifically, at each decoding step $j$, we compute the next-token distributions given the prefix $a_{<j}$:
\begin{align}
p_{\texttt{par}}
&= P(a_j \mid q, a_{<j}; \mathcal{M}_u^{(t)})
\label{eq:p_par}\\
p_{\texttt{ret}}
&= P(a_j \mid q, \mathbf{c}_{\texttt{ret}}, a_{<j}; \mathcal{M}_u^{(t)})
\label{eq:p_ret}
\end{align}

We then fuse these distributions via probability interpolation with a mixing coefficient $\lambda$:
\begin{equation}
p_{\texttt{fin}}
= \lambda \cdot p_{\texttt{ret}}
+ (1-\lambda) \cdot p_{\texttt{par}}
\label{eq:interp}
\end{equation}

Finally, we select the next token as
$\hat{a}_j = \arg\max_a p_{\texttt{fin}}(a)$
and append it to the shared prefix $a_{<j}$ for the next decoding step.

\section{Experiments}\label{sec:4}

\subsection{Setups}\label{sec:4.1}

\paragraph{Datasets and metrics.}
We evaluate \name{} on the \textbf{LongLaMP} \citep{kumar2024longlamp} benchmark, a long-form personalized text generation benchmark built from real-user data. 
Specifically, Abstract Generation and Review Writing are selected for their accessible temporal metadata, unlike Topic Writing (no timestamps) and Email Generation (restricted access).
Both tasks consist of naturally occurring histories from real researchers and consumers.
For the Abstract Generation task, we chronologically sort each user's samples by publication \texttt{year}. 
However, for Review Writing, the LongLaMP dataset does not provide timestamps directly, so we recover the \texttt{reviewTime} by matching each instance to the original Amazon Product Review dataset \citep{ni2019justifying} and sort them chronologically.
After temporal ordering, we retain users with sufficient interaction history for longitudinal analysis, resulting in 100 users per task.
We further divide these users into \textit{Small} and \textit{Large} groups to evaluate performance under different history-length regimes.
To simulate realistic continual personalization, we partition each user's chronologically ordered interaction history into five consecutive periods, denoted as Period 0--4.
In this setup, Period 0 is used as the initial observed history for personalization, while Periods 1--4 are revealed sequentially as new user data arrives.
Within each period, we split data chronologically, where the earliest 90\% of samples are treated as observed data available for updating, and the most recent 10\% are held out for evaluation.

To verify that the periodization captures meaningful non-stationarity, we characterize longitudinal changes over the full set of users across complementary dimensions.

\begin{table}[t]
\centering
\small
\caption{\textbf{Temporal drift across user histories.}
Comparison of temporal drift across feature dimensions and periods against a random within-user baseline for Abstract Generation and Review Writing. Drift Ratio denotes the ratio of P0$\rightarrow$P4 drift to the random within-user baseline.}
\label{tab:temporal_drift}

\resizebox{\columnwidth}{!}{%
\begin{tabular}{l|cccc}
\toprule[1.5pt]

\multirow{2}{*}[-2pt]{\textbf{Dimension}}
& \multicolumn{4}{c}{\textbf{Abstract Generation}} \\
\cmidrule(lr){2-5}
& Random
& P0$\rightarrow$P1
& P0$\rightarrow$P4
& Ratio \\

\midrule[1pt]

Semantic        & 0.039 & 0.104 & 0.278 & 7.13$\times$ \\
Topic           & 0.182 & 0.254 & 0.338 & 1.86$\times$ \\
Lexical/content & 0.087 & 0.136 & 0.178 & 2.04$\times$ \\
Style           & 0.002 & 0.004 & 0.007 & 2.79$\times$ \\

\midrule[1.5pt]

\multirow{2}{*}[-2pt]{\textbf{Dimension}}
& \multicolumn{4}{c}{\textbf{Review Writing}} \\
\cmidrule(lr){2-5}
& Random
& P0$\rightarrow$P1
& P0$\rightarrow$P4
& Ratio \\

\midrule[1pt]

Semantic            & 0.038 & 0.100 & 0.158 & 4.11$\times$ \\
Lexical/content     & 0.072 & 0.132 & 0.175 & 2.42$\times$ \\
Sentiment           & 0.080 & 0.099 & 0.178 & 2.22$\times$ \\
Preference polarity & 0.015 & 0.024 & 0.040 & 2.69$\times$ \\
Style               & 0.002 & 0.007 & 0.011 & 4.33$\times$ \\

\bottomrule[1.5pt]
\end{tabular}%
}

\vspace{-0.1in}
\end{table}
\begin{itemize}[leftmargin=3.5mm, topsep=2pt, itemsep=2pt, parsep=0pt]
    \item[$\circ$] \textbf{Abstract Generation.}
    We use centroid cosine distance for \textit{semantic drift} and keyword, bigram, and function-word JSD for \textit{topic}, \textit{lexical/content}, and \textit{style shifts}, respectively \citep{lin1991divergence, kestemont2014function}.

    \item[$\circ$] \textbf{Review Writing.}
    We use the same \textit{semantic}, \textit{lexical/content}, and \textit{style} metrics, with Wasserstein distance for \textit{sentiment} \citep{schmitz2018wasserstein} and rating JSD for \textit{preference polarity}.
\end{itemize}
We compare Period 0 (P0) $\rightarrow$ Period 1 (P1) and P0 $\rightarrow$ Period 4 (P4) to capture short- and long-term divergence from the initial user state, using random within-user variation as a non-temporal baseline.
Table~\ref{tab:temporal_drift} shows that P0 $\rightarrow$ P4 drift exceeds the random baseline in every dimension, by 1.86--7.13$\times$ for Abstract Generation and 2.22--4.33$\times$ for Review Writing, and is consistently greater than P0 $\rightarrow$ P1 drift.
These full-set, method-independent patterns indicate multidimensional non-stationarity in the evaluation streams, without assuming that every temporal difference directly represents a preference change.

To evaluate performance, we employ ROUGE-1 and ROUGE-L to measure content fidelity and style alignment with the ground truth.
We present the results for each individual period as well as the average performance (Period Avg) to show the overall trends.

\paragraph{Baselines.}
We compare \name{} against several baselines categorized as follows: (1) \textit{Base} -- generating responses using the vanilla LLM without any personalization or fine-tuning.
For \textbf{standard personalization} baselines (2--4), we evaluate two setups: \textit{each} (using only current period data) and \textit{accum} (using cumulative data from period 0 to current period). 
These include: 
(2) \textit{RAG} \citep{hwang2023aligning} -- augmenting input with retrieved history from the respective corpus; 
(3) \textit{PAG} \citep{richardson2023integrating} -- utilizing summarized user profiles generated from the corresponding data scope; 
and (4) \textit{OPPU} \citep{tan2024democratizing} -- fine-tuning user-specific LoRA adapters \citep{hu2022lora} either locally on the current period or sequentially on the cumulative stream. 
We also include \textbf{continual learning} baselines: 
(5) \textit{CAMA} \citep{kim2024online} -- a replay-based method that selectively buffers high-confidence samples exceeding the previous period's average confidence and mixes them with historical data for training; 
(6) \textit{SAR} \citep{seo2024budgeted} -- a method that prioritizes replay samples from under-represented clusters based on effective use-frequency to mitigate overfitting to dominant patterns; 
(7) \textit{O-LoRA} \citep{wang2023orthogonal} -- a method that projects new parameter updates onto a subspace orthogonal to previous adapters to reduce interference; 
and (8) \textit{EWC-LoRA} \citep{xiang2023language} -- a method that applies elastic weight consolidation with Fisher Information to penalize drastic changes in parameters critical for prior periods. 

\paragraph{Implementation details.} 
In our experiments, we utilize \texttt{Gemma-3-4B-IT} \citep{team2025gemma} as the backbone LLM.
For training, we employ AdamW \citep{loshchilov2017decoupled} with a weight decay of 0.01 and a learning rate of $1 \times 10^{-4}$, utilizing a linear decay scheduler with a warmup ratio of 0.1.
We use a per-device batch size of 4, train for 2 epochs per period, and truncate the maximum sequence length to 768 tokens.
For LoRA, we target the query and value projection layers with a rank $r=8$, a LoRA scaling factor of 8, dropout of 0.05, and no bias. 
During inference, we ensure deterministic outputs using greedy decoding with a repetition penalty of 1.2, setting the maximum generated tokens to 600. 

Regarding the hyperparameters for \name{}, we set the balancing term $\alpha$ in the drift score function (Eq.~\ref{eq:drift_score}) to 1.0 and construct the drift set by retaining the top 30\% ($p=30$) of samples. 
For the residual replay buffer, we maintain a maximum budget of $N_{\max} = 50$ samples per user.
During inference, we retrieve $k=2$ documents using BM25 \citep{robertson2004simple} as \textsc{Sim} and apply a relevance gating threshold $\sigma$, defined as the task-specific third quartile (Q3) of the retrieval score distribution aggregated across all users and periods, to filter out less relevant contexts.
Finally, we set the mixing coefficient $\lambda$ for probability interpolation (Eq.~\ref{eq:interp}) to 0.5 to balance the parametric and non-parametric distributions equally.
\begin{table*}[t]
\caption{\textbf{Main results.} ROUGE-1 and ROUGE-L scores across periods (1--4) and their average for Abstract Generation and Review Writing tasks. All experiments are conducted using \texttt{Gemma-3-4B-IT}. \textbf{Bold} indicates the best performance, and \underline{underline} indicates the second best.}
\label{tab:main}
\centering

\setlength{\tabcolsep}{3pt}
\renewcommand{\arraystretch}{1.05}

\resizebox{\textwidth}{!}{%
\begin{tabular}{l|cccccccc|cc}
\toprule[1.5pt]

& \multicolumn{10}{c}{\textbf{Abstract Generation}} \\ 
\cmidrule(lr){2-11} 

& \multicolumn{2}{c}{\textbf{Period 1}}
& \multicolumn{2}{c}{\textbf{Period 2}}
& \multicolumn{2}{c}{\textbf{Period 3}}
& \multicolumn{2}{c|}{\textbf{Period 4}}
& \multicolumn{2}{c}{\textbf{Period Avg}} \\ 
\cmidrule(lr){2-3} \cmidrule(lr){4-5} \cmidrule(lr){6-7} \cmidrule(lr){8-9} \cmidrule(lr){10-11}

\multirow{-3}{*}{\textbf{Method}} & ROUGE-1 & ROUGE-L
& ROUGE-1 & ROUGE-L
& ROUGE-1 & ROUGE-L
& ROUGE-1 & ROUGE-L
& ROUGE-1 & ROUGE-L \\ 

\midrule[1pt]

Base       & 0.302 & 0.147 & 0.302 & 0.148 & 0.300 & 0.148 & 0.298 & 0.144 & 0.301 & 0.147 \\
\midrule
RAG each   & 0.304 & 0.162 & 0.302 & 0.160 & 0.301 & 0.160 & 0.299 & 0.156 & 0.301 & 0.159 \\
RAG accum  & 0.301 & 0.161 & 0.300 & 0.160 & 0.297 & 0.158 & 0.296 & 0.155 & 0.299 & 0.158 \\
PAG each   & 0.304 & 0.144 & 0.303 & 0.145 & 0.299 & 0.144 & 0.304 & 0.144 & 0.302 & 0.144 \\
PAG accum  & 0.306 & 0.145 & 0.303 & 0.144 & 0.303 & 0.145 & 0.307 & 0.146 & 0.305 & 0.145 \\
OPPU each  & 0.321 & 0.155 & 0.320 & 0.155 & 0.316 & 0.153 & 0.318 & 0.150 & 0.319 & 0.153 \\
OPPU accum & 0.319 & 0.154 & 0.305 & 0.151 & 0.292 & 0.144 & 0.282 & 0.136 & 0.300 & 0.146 \\ 
\midrule 
CAMA       & 0.336 & 0.162 & \underline{0.338} & \underline{0.166} & \underline{0.337} & \underline{0.166} & \underline{0.337} & \underline{0.162} & \underline{0.337} & \underline{0.164} \\
SAR        & 0.333 & \underline{0.163} & 0.334 & 0.164 & 0.324 & 0.160 & 0.314 & 0.151 & 0.326 & 0.159 \\
O-LoRA     & 0.334 & \underline{0.163} & 0.303 & 0.148 & 0.300 & 0.147 & 0.299 & 0.145 & 0.309 & 0.151 \\
EWC-LoRA   & \underline{0.337} & \underline{0.163} & 0.335 & 0.165 & 0.328 & 0.163 & 0.323 & 0.157 & 0.331 & 0.162 \\ 
\midrule
\rowcolor[HTML]{D9D9D9}
\name{} & \textbf{0.346} & \textbf{0.188} & \textbf{0.353} & \textbf{0.192} & \textbf{0.351} & \textbf{0.191} & \textbf{0.351} & \textbf{0.190} & \textbf{0.350} & \textbf{0.190} \\

\midrule[1.5pt] 

& \multicolumn{10}{c}{\textbf{Review Writing}} \\ 
\cmidrule(lr){2-11}

& \multicolumn{2}{c}{\textbf{Period 1}}
& \multicolumn{2}{c}{\textbf{Period 2}}
& \multicolumn{2}{c}{\textbf{Period 3}}
& \multicolumn{2}{c|}{\textbf{Period 4}}
& \multicolumn{2}{c}{\textbf{Period Avg}} \\ 
\cmidrule(lr){2-3} \cmidrule(lr){4-5} \cmidrule(lr){6-7} \cmidrule(lr){8-9} \cmidrule(lr){10-11}

\multirow{-3}{*}{\textbf{Method}} & ROUGE-1 & ROUGE-L
& ROUGE-1 & ROUGE-L
& ROUGE-1 & ROUGE-L
& ROUGE-1 & ROUGE-L
& ROUGE-1 & ROUGE-L \\ 

\midrule[1pt]

Base       & 0.301 & \underline{0.125} & 0.302 & 0.125 & 0.303 & 0.125 & 0.303 & 0.125 & 0.302 & \underline{0.125} \\
\midrule
RAG each   & 0.273 & 0.122 & 0.275 & 0.124 & 0.278 & 0.123 & 0.279 & 0.125 & 0.276 & 0.123 \\
RAG accum  & 0.274 & 0.122 & 0.272 & 0.122 & 0.277 & 0.124 & 0.278 & 0.125 & 0.275 & 0.123 \\
PAG each   & 0.289 & 0.120 & 0.290 & 0.122 & 0.285 & 0.121 & 0.281 & 0.120 & 0.286 & 0.121 \\
PAG accum  & 0.286 & 0.120 & 0.288 & 0.120 & 0.288 & 0.121 & 0.279 & 0.120 & 0.285 & 0.120 \\
OPPU each  & 0.304 & 0.124 & 0.307 & 0.125 & 0.310 & \underline{0.126} & 0.307 & 0.125 & 0.307 & \underline{0.125} \\
OPPU accum & \underline{0.310} & 0.123 & 0.312 & 0.123 & 0.308 & 0.122 & 0.306 & 0.123 & 0.309 & 0.123 \\ 
\midrule
CAMA       & 0.305 & \underline{0.125} & 0.313 & \underline{0.126} & 0.317 & 0.125 & \underline{0.314} & 0.124 & \underline{0.312} & \underline{0.125} \\
SAR        & 0.309 & 0.124 & \underline{0.316} & 0.124 & 0.314 & 0.124 & 0.308 & 0.123 & \underline{0.312} & 0.124 \\
O-LoRA     & 0.306 & \underline{0.125} & 0.301 & 0.125 & 0.304 & 0.125 & 0.304 & \underline{0.126} & 0.304 & \underline{0.125} \\
EWC-LoRA   & 0.304 & 0.124 & 0.315 & 0.125 & \underline{0.319} & 0.124 & 0.312 & 0.124 & \underline{0.312} & 0.124 \\ 
\midrule
\rowcolor[HTML]{D9D9D9}
\name{} & \textbf{0.325} & \textbf{0.141} & \textbf{0.326} & \textbf{0.142} & \textbf{0.329} & \textbf{0.143} & \textbf{0.324} & \textbf{0.142} & \textbf{0.326} & \textbf{0.142} \\

\bottomrule[1.5pt]
\end{tabular}%
}
\vspace{-0.15in}
\end{table*}

\subsection{Main results}
Table~\ref{tab:main} presents the evaluation results of \name{} against baselines on Abstract Generation and Review Writing.
First, \name{} consistently achieves the best performance across all evaluation periods for both tasks, with gains both in ROUGE-1 and ROUGE-L metrics. 
To be specific, in Abstract Generation, our method achieves relative improvements of 3.86\% in ROUGE-1 and 15.85\% in ROUGE-L over the strongest CL baseline.
Similarly, in the Review Writing task, \name{} maintains a lead over existing competitive approaches, achieving improvements of 4.49\% in ROUGE-1 and 13.60\% in ROUGE-L. 
{\looseness=-1 These results indicate that our proposed approach effectively balances plasticity and stability, delivering substantial performance gains over the strongest existing methods.\par}

Furthermore, we observe that standard personalization approaches (\textit{e.g.}, RAG, PAG) can be ineffective or even detrimental in certain contexts.
Notably, in the Review Writing task, both the Each and Accum variants of RAG and PAG fail to surpass the non-personalized Base model.
For instance, RAG each results in an 8.60\% drop in ROUGE-1 compared to the Base model, indicating a vulnerability to noisy or irrelevant context.
However, \name{} overcomes this limitation, demonstrating an 18.12\% relative improvement over RAG each.
This suggests that \name{} selectively leverages beneficial context while filtering irrelevant history. 

Finally, CL baselines show greater stability than the naive accumulation strategy.
We observe a general trend where replay-based approaches (\textit{e.g.}, CAMA, SAR) tend to yield robust performance, slightly outperforming or matching regularization-based methods like O-LoRA. 
However, even these strong baselines fall short of \name{}.
This limitation likely stems from treating historical samples uniformly or relying on fixed heuristics, which are not designed to prioritize drift-relevant update. 
In contrast, \name{} achieves superior results by explicitly targeting and adapting to meaningful preference drifts, allowing for more efficient and effective personalization.\looseness=-1
Additional validation and LLM generalization results are provided in Appendix Sections~\ref{sec:additional_evaluation} and~\ref{sec:generalization_llms}.\looseness=-1

\subsection{Further Analyses}

\paragraph{Component-wise ablation.}
Table~\ref{tab:component_ablation} presents a unified component-wise ablation, showing that the gains from \name{} emerge progressively rather than from a single component.
First, adding the user adapter improves both tasks, yielding relative gains of 7.6\% in ROUGE-1 and 5.4\% in ROUGE-L on Abstract Generation, and 3.6\% in ROUGE-1 on Review Writing (2nd row). 
This confirms that parametric personalization provides an independent benefit.
Next, adding retrieval-interpolated generation yields the largest gain, especially on Abstract Generation, where it improves ROUGE-1 by 8.3\% and ROUGE-L by 23.2\% over the adapter-only setting (3rd row). 
This suggests that retrieved evidence complements what is internalized in the adapter and recovers information that parametric updates alone may miss.
Lastly, relevance gating then provides a further 2.2\% gain in ROUGE-1 on Review Writing while leaving Abstract Generation nearly unchanged, suggesting that it mainly helps by suppressing off-topic retrieval when retrieved context is noisy (4th row).
We next examine the training-side selection and inference-side gating mechanisms separately. 
\begin{table}[t]
    \centering
    \caption{\textbf{Component-wise ablation.} Impact of progressively adding adapter-based updating (Eq.~\ref{eq:drift_score}), retrieval-interpolated generation (Eq.~\ref{eq:interp}), and relevance gating (Eq.~\ref{eq:gating}) on Abstract Generation (\textbf{Abs}) and Review Writing (\textbf{Rev}). R-1 and R-L represent ROUGE-1 and ROUGE-L, respectively.}
    \label{tab:component_ablation}
    
    \resizebox{\columnwidth}{!}{%
        \begin{tabular}{l|ccc|cc|cc}
            \toprule[1.5pt]
            \multirow{2}{*}{\textbf{Method}} 
            & \multirow{2}{*}{Adapter} 
            & \multirow{2}{*}{Interp.} 
            & \multirow{2}{*}{Gating} 
            & \multicolumn{2}{c|}{\textbf{Abs}} 
            & \multicolumn{2}{c}{\textbf{Rev}} \\
            \cmidrule(lr){5-6} \cmidrule(lr){7-8}
            & & & & R-1 & R-L & R-1 & R-L \\
            \midrule[1pt]
            Base            & \xmark & \xmark & \xmark & 0.301 & 0.147 & 0.302 & 0.125 \\
                        & \cmark & \xmark & \xmark & 0.324 & 0.155 & 0.313 & 0.125 \\
               & \cmark & \cmark & \xmark & 0.351 & 0.191 & 0.319 & 0.143 \\
            \midrule
            \rowcolor[HTML]{D9D9D9}
            \name{}   & \cmark & \cmark & \cmark & 0.350 & 0.190 & 0.326 & 0.142 \\
            \bottomrule[1.5pt]
        \end{tabular}%
    }
\end{table}

\begin{table}[t]
    \centering
\caption{\textbf{Effectiveness of selective adaptation.} Performance comparison across data selection criteria for period-wise adapter updates.}
    \label{tab:train_ablation}

    \resizebox{\columnwidth}{!}{%
        \begin{tabular}{l|cc|cc}
            \toprule[1.5pt]

            \multirow{2}{*}{\textbf{Selection}}
            & \multicolumn{2}{c|}{\textbf{Abstract Generation}} & \multicolumn{2}{c}{\textbf{Review Writing}} \\

            \cmidrule(lr){2-3} \cmidrule(lr){4-5}
            & ROUGE-1 & ROUGE-L & ROUGE-1 & ROUGE-L \\

            \midrule[1pt]

            All-100\%       & 0.300 & 0.146 & 0.309 & 0.123 \\
            Bottom-30\%     & 0.284 & 0.142 & 0.309 & 0.124 \\ 
            Random-30\%     & 0.312 & 0.152 & 0.312 & 0.125 \\ 
            \rowcolor[HTML]{D9D9D9}
            Top-30\% (Ours) & 0.324 & 0.155 & 0.313 & 0.125 \\

            \bottomrule[1.5pt]
        \end{tabular}%
    }
    \vspace{-0.1in}
\end{table}
\paragraph{Selective adaptation.}
In our method, the adapter is updated at each period using only the 30\% of interactions with the highest drift scores.
These high-scoring interactions reflect patterns supported by the base model but not yet internalized by the current user adapter.
To evaluate this ranking, Table~\ref{tab:train_ablation} compares \textit{Top-30\%} with \textit{All-100\%}, an equally sized random subset (\textit{Random-30\%}), and the lowest-scoring subset (\textit{Bottom-30\%}).
Despite using 70\% fewer update samples, \textit{Top-30\%} outperforms \textit{All-100\%} on both tasks and matches or improves upon \textit{Random-30\%} across all task--metric pairs.
The difference is particularly clear in Abstract Generation, where it improves ROUGE-1 by 8.0\% over \textit{All-100\%} and by 14.1\% over \textit{Bottom-30\%}.
These results indicate that the drift score prioritizes interactions with greater adaptation value, while indiscriminate or low-score updates can dilute useful signals.
Further analyses of the selection ratio and drift-score formulation are provided in Appendix~\ref{sec:training_analyses}.

\begin{table}[t]
    \centering
    \caption{\textbf{Effectiveness of relevance gating.} Performance comparison across BM25 score thresholds based on BM25 score quartiles. \textit{Interp. Ratio} denotes the percentage of queries that activate the retrieval-interpolated generation branch.}
    \label{tab:threshold}

    \resizebox{\columnwidth}{!}{%
        \begin{tabular}{l|ccc|ccc}
            \toprule[1.5pt]
            \multirow{2}{*}{\textbf{Threshold}}
            & \multicolumn{3}{c|}{\textbf{Abstract Generation}}
            & \multicolumn{3}{c}{\textbf{Review Writing}} \\
            \cmidrule(lr){2-4} \cmidrule(lr){5-7}
            & \shortstack[c]{Interp.\\Ratio (\%)} 
            & \raisebox{4pt}[0pt][0pt]{R-1} 
            & \raisebox{4pt}[0pt][0pt]{R-L}
            & \shortstack[c]{Interp.\\Ratio (\%)} 
            & \raisebox{4pt}[0pt][0pt]{R-1} 
            & \raisebox{4pt}[0pt][0pt]{R-L} \\            
            \midrule[1pt]
            $\sigma=0$                & 100.0 & 0.351 & 0.191 & 100.0 & 0.319 & 0.143 \\
            $\sigma=\text{Q2}$        & 97.4  & 0.350 & 0.190 & 65.5  & 0.323 & 0.142 \\
            \rowcolor[HTML]{D9D9D9}
            $\sigma=\text{Q3}$ (Ours) & 92.3  & 0.350 & 0.190 & 41.4  & 0.326 & 0.142 \\
            \bottomrule[1.5pt]
        \end{tabular}%
    }
    \vspace{-0.15in}
\end{table}
\paragraph{Relevance gating analysis.}
We further analyze relevance gating to understand when retrieval should be activated and how this choice differs across tasks (Table~\ref{tab:threshold}). 
In Abstract Generation, semantic continuity is strong, so over 92\% of queries trigger retrieval even under a strict threshold, suggesting that retrieved contexts are typically relevant.
In contrast, Review Writing has more diverse content and retrieved histories are frequently off topic, making indiscriminate retrieval harmful. 
Under Q3, gating filters out 58.6\% of queries that fall below the cutoff, yet still yields a 2.19\% relative gain in ROUGE-1. 
This supports that gating removes interfering context and activates retrieval only when there is sufficient evidence. 
It also provides a practical quality--cost control knob. 
Under Q3, only 41.4\% of Review Writing queries trigger retrieval, and the remaining queries fall back to single-path decoding, reducing inference-time compute compared to always running dual-path interpolation. 
Further analyses of relevance thresholds and inference-time cost optimization are provided in Appendix Sections~\ref{sec:appendix_inf_overhead} and~\ref{sec:appendix_optim}.

\begin{figure*}[!t]
\centering
\includegraphics[width=1.0\textwidth]{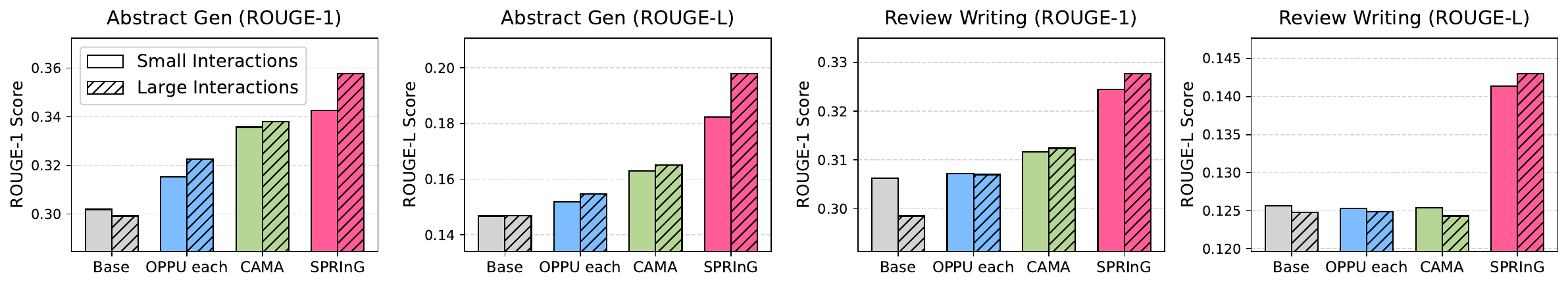}
\caption{
\textbf{Interaction scalability of \name{}.} Performance comparison on different interaction history lengths.
}
\vspace{-0.1in}
\label{fig:robustness}
\end{figure*}
\paragraph{Robustness across the number of interactions.}
To investigate the adaptability of \name{} to varying scales of user data, we categorized users into two groups: \textit{Small Interactions} and \textit{Large Interactions}. 
Definitions and interaction statistics for these groups are provided in Appendix~\ref{supp:profile_size} and Table~\ref{tab:interaction_stats}.
The \textit{Small Interactions} group represents a scenario of relative data sparsity, whereas the \textit{Large Interactions} group includes users with extensive histories, presenting challenges related to noise accumulation and retrieval efficiency.
Figure~\ref{fig:robustness} illustrates the performance comparison across these groups.
In Abstract Generation, \name{} demonstrates robust scalability, consistently outperforming baselines with large gains in both settings.
In Review Writing, baselines such as Base and OPPU exhibit performance degradation within the \textit{Large Interactions} group, likely due to noise accumulation from extended histories. 
In contrast, \name{} effectively leverages the extended contexts to improve performance, validating the efficacy of our selective mechanism in filtering noise and extracting relevant signals.

\paragraph{Additional analyses in the Appendix.}
Due to space limits, several complementary analyses are deferred to the Appendix.
These include component scaling, memorization-risk checks, robustness under abrupt preference shifts, evaluation on an additional dataset, and cost analyses (Appendix Sections~\ref{sec:component_scaling}--\ref{sec:appendix_cost_analysis}).
\section{Related Work}

\noindent\textbf{LLM personalization.} 
LLM personalization aligns model behaviors with individual preferences \citep{zhang2024personalization}, broadly categorized into \textit{non-parametric} and \textit{parametric} approaches.
Non-parametric methods condition generation on user context via retrieval of historical data \citep{hwang2023aligning} or summarized profiles \citep{richardson2023integrating}.
While interpretable and easy to deploy, their effectiveness relies on retrieval quality and context window constraints.
Conversely, parametric methods directly encode preferences by fine-tuning user-specific parameters, often employing efficient adapters like LoRA \citep{tan2024democratizing} or hierarchical structures \citep{zhang2025proper}.
However, these approaches primarily focus on static optimization, overlooking the temporal evolution of user traits required for \textit{continual} personalization.

\paragraph{Continual learning.} 
Continual learning (CL) addresses the challenge of sequentially updating models while mitigating catastrophic forgetting \citep{ke2022continual, shi2024continual}.
Research typically focuses on two complementary strategies: \textit{regularization} and \textit{replay}.
Regularization methods constrain the optimization trajectory to preserve prior knowledge, such as applying Fisher-weighted penalties \citep{kirkpatrick2017overcoming} or enforcing orthogonality in subspace updates \citep{wang2023orthogonal}.
Conversely, replay-based methods maintain a buffer to interleave past examples with current streams.
To maximize efficiency, recent work prioritizes \textit{selective} data retention over uniform sampling \citep{rolnick2019experience}, utilizing metrics like interference \citep{aljundi2019online}, confidence scores \citep{kim2024online}, activation magnitudes \citep{du2024unlocking}, or spaced repetition heuristics \citep{m2024leitner} to identify the most informative samples.
Unlike prior selection strategies that primarily determine which past examples to retain for replay, our drift score identifies current interactions that remain insufficiently captured by the user adapter and prioritizes them for adaptation.
\section{Conclusion}
We propose \name{}, a semi-parametric framework for continual LLM personalization under evolving user preferences.
Our method addresses the limitations of CL and retrieval-based approaches through (1) drift-driven selective adaptation and (2) gated retrieval-interpolated generation.
First, temporal analysis validates our non-stationary setup by revealing multidimensional shifts beyond random within-user variation that grow over time.
Under these evolving user streams, \name{} consistently outperforms standard personalization and CL baselines across both tasks and periods.
Component analyses show that selective adapter updates, retrieval-interpolated generation, and relevance gating provide complementary benefits by prioritizing informative changes, recovering residual evidence, and suppressing irrelevant context.
Finally, robustness across interaction scales demonstrates \name{}'s effectiveness under varying data availability.
Collectively, our findings underscore the importance of distinguishing genuine preference drift and suggest a scalable path for deploying long-term personalized LLMs in practice.

\newpage
\section*{Limitations} 
While the proposed approach yields substantial gains, it comes with practical trade-offs that offer directions for future research.
First, the dual-path probability interpolation strategy introduces inference latency and cost overhead, as it requires two forward passes per decoding step.
However, this is partially offset by our selective training strategy, which updates parameters on only 30\% of the data, reducing the overall training cost compared to exhaustive retraining.
To further optimize inference efficiency, future work could explore adaptive interpolation strategies, such as terminating the dual-path process mid-generation or reverting to a single parametric path once the divergence between the two distributions falls below a certain threshold.
As discussed in Appendix~\ref{sec:appendix_inf_overhead} and Appendix~\ref{sec:appendix_optim}, this overhead is often smaller in the gated retrieval setup, and we additionally outline several optimization strategies to reduce dual-path computation.

Next, the reliability of the drift score may be challenged by transient noise or abrupt topic shifts (false drift) that do not reflect persistent preference evolution.
In our framework, we reduce the impact of such noise by restricting updates to a small, high-novelty subset and preserving hard-but-informative cases in a bounded residual buffer, while inference-time relevance gating additionally prevents noisy retrieved histories from interfering with generation.
Appendix~\ref{sec:diverse_dataset} and Appendix~\ref{sec:abrupt_shift} provide further analysis of this issue under noisier user streams and abrupt-shift scenarios, suggesting that the current framework is reasonably robust to both conditions.
Future research could enhance this by adopting more sophisticated retrieval and gating frameworks, such as dense vector-based retrieval or cross-encoder re-ranking, to more precisely isolate semantically relevant evidence from diverse interaction streams.

Lastly, scaling user-specific adapters poses an infrastructure challenge as the user base grows.
Nevertheless, the parameter-efficient nature of LoRA and modern multi-adapter serving techniques make this approach increasingly viable for large-scale deployment.
As a potential extension, exploring hierarchical or shared adapter architectures that can group users with similar preference trajectories could alleviate storage burdens while facilitating cross-user knowledge transfer \citep{tan2024democratizing, zhang2025proper}.

\section*{Broader Impact and Ethical Implications}
The deployment of \name{} for continual personalization can improve long-term user experience and reduce unnecessary computation by avoiding indiscriminate retraining or unfiltered retrieval, thereby enabling more resource-efficient adaptation.
However, strong personalization may reinforce biases and create echo chambers by repeatedly optimizing for a user's current viewpoint; practical deployments should include mechanisms that promote diversity and provide transparent user controls (\textit{e.g.,} reset or edit preferences).
In addition, responsive adaptation raises safety risks, including preference poisoning attacks and potential drift toward harmful objectives.
Future systems should therefore integrate safety-aware filtering and anomaly detection to prevent malicious or unsafe signals from being incorporated into personalization updates.

\section*{Acknowledgments}

All authors are affiliated with the Department of Artificial Intelligence at Yonsei University.
This research was supported in part by the Institute for Information \& Communications Technology Planning \& Evaluation (IITP) grant funded by the Korea government (MSIT) (No. RS-2020-II201361, Artificial Intelligence Graduate School Program (Yonsei University); No. RS-2025-25442405, Development of a Self-Learning World Model-Based
AGI System for Hyperspectral Imaging).
\bibliography{custom}
\appendix
\newpage
\begin{algorithm}[H]
   \caption{\name{} training algorithm}
   \label{alg:training}
\begin{algorithmic}
  \State
  \textbf{Input:} Base LLM $\mathcal{M}_\theta$ (frozen), initial user adapter $\phi_u^{(0)}$, interaction stream $\{\mathcal{D}_u^{(t)}\}_{t\ge1}$,
  scoring function $\mathcal{S}(\cdot)$ (Eq.~\ref{eq:drift_score}), selection ratio $p$, buffer budget $N_{\max}$
  \vspace{0.05in}
  \hrule
  \vspace{0.05in}

  \State $\mathcal{B}_u^{(0)} \leftarrow \emptyset$
  \For{$t=1,2,\cdots$}
    \State \texttt{\color{Gray2} /* Drift sample selection */ }
    \For{$(q,a)\in \mathcal{D}_u^{(t)}$}
      \State $\mathcal{S}_t(q,a) \leftarrow \mathcal{S}(q,a;\mathcal{M}_\theta,\phi_u^{(t-1)})$
    \EndFor
    \State $\tau_p \leftarrow$ top-$p\%$ threshold on $\mathcal{D}_u^{(t)}$
    \State $\mathcal{D}_{\texttt{drift}}^{(t)} \leftarrow \{(q,a)\in \mathcal{D}_u^{(t)} ~|~ \mathcal{S}_t(q,a)\ge \tau_p\}$
    \Statex \vspace{-0.6\baselineskip}
    \State \texttt{\color{Gray2} /* Parametric adapter update */ }
    \State $\phi_u^{(t)} \leftarrow \phi_u^{(t-1)} \oplus \mathcal{D}_{\texttt{drift}}^{(t)}$
    \Statex \vspace{-0.6\baselineskip}
    \State \texttt{\color{Gray2} /* Replay buffer construction */ }
    \State $\mathcal{C}_u^{(t)} \leftarrow \mathcal{D}_{\texttt{drift}}^{(t)} \cup \mathcal{B}_u^{(t-1)}$
    \For{$(q,a)\in \mathcal{C}_u^{(t)}$}
      \State $\mathcal{S}'_t(q,a) \leftarrow \mathcal{S}(q,a;\mathcal{M}_\theta,\phi_u^{(t)})$
    \EndFor    
\State If $|\mathcal{C}_u^{(t)}| \le N_{\max}$,
$\mathcal{B}_u^{(t)} \leftarrow \mathcal{C}_u^{(t)}$; otherwise,
    \State $\mathcal{B}_u^{(t)} \leftarrow
    \operatorname{TopK}_{N_{\max}}
    (\mathcal{C}_u^{(t)};\mathcal{S}'_t)$
  \EndFor

\State \textbf{return} period-wise adapters $\{\phi_u^{(t)}\}$ and replay buffers $\{\mathcal{B}_u^{(t)}\}$
\end{algorithmic}
\end{algorithm}
\vspace{-0.15in}

\begin{algorithm}[H]
   \caption{\name{} inference algorithm}
   \label{alg:inference}
\begin{algorithmic}
  \State
  \textbf{Input:} User-adapted model $\mathcal{M}_u^{(t)}$, replay buffer $\mathcal{B}_u^{(t)}$,
  test query $q$, retrieval top-$k$, relevance threshold $\sigma$, interpolation weight $\lambda$
  \vspace{0.05in}
  \hrule
  \vspace{0.05in}

\State \texttt{\color{Gray2} /* Relevance-gated retrieval */ }
    \State $\mathcal{R}_q \leftarrow
    \operatorname{TopK}_{k}
    \left(\mathcal{B}_u^{(t)}; \textsc{Sim}(q,\cdot)\right)$
    \State $\mathbf{c}_{\texttt{ret}} \leftarrow
    \textsc{Concat}
    \left\{
    h \in \mathcal{R}_q
    \mid
    \textsc{Sim}(q,h) > \sigma
    \right\}$
\Statex \vspace{-0.6\baselineskip}

  \State \texttt{\color{Gray2} /* Probability interpolation */ }
  \State $a_{<1} \leftarrow \emptyset$
\For{$j=1,2,\cdots$ \textbf{until} \textsc{EOS}}
    \State Compute $p_{\texttt{par}}$ (Eq.~\ref{eq:p_par})
    \If{$\mathbf{c}_{\texttt{ret}} \neq \emptyset$}
        \State Compute $p_{\texttt{ret}}$ (Eq.~\ref{eq:p_ret})
        \State Compute $p_{\texttt{fin}}$ by interpolation (Eq.~\ref{eq:interp})
    \Else
        \State $p_{\texttt{fin}} \leftarrow p_{\texttt{par}}$
    \EndIf
    \State $\hat{a}_j \leftarrow \arg\max_a p_{\texttt{fin}}(a)$
    \State $a_{<j+1} \leftarrow a_{<j} \oplus \hat{a}_j$
\EndFor
  \State \textbf{return} generated response $a$
\end{algorithmic}
\end{algorithm}

\section{Training and Inference Algorithms}

Algorithm~\ref{alg:training} presents the training pipeline for \name{}.
It identifies drift samples based on the scoring function (Eq.~\ref{eq:drift_score}), updates the parametric adapter, and maintains the replay buffer within the budget $N_{\max}$.
The likelihood terms in the drift score are computed using mean token log-likelihood under the same maximum sequence length for both the base and adapted models.

Algorithm~\ref{alg:inference} details the inference process.
It incorporates relevance-gated retrieval and probability interpolation (Eq.~\ref{eq:interp}) to generate the final response.

\begin{table*}[t]
    \centering
    \small
    \renewcommand{\arraystretch}{1.2} 
    \setlength{\tabcolsep}{0pt}
    
    \caption{\textbf{Detailed statistics regarding the number of interactions.} 
    Comparison between Small and Large interaction groups across tasks.}
    \label{tab:interaction_stats}
    
    \begin{tabular*}{0.95\textwidth}{@{\extracolsep{\fill}}llcccc}
        \toprule
        \textbf{Task} & \textbf{Group} & \textbf{N} & \textbf{Mean \# Interactions} & \textbf{Median} & \textbf{Range (Min--Max)} \\
        \midrule
        \multirow{2}{*}{Abstract Generation} & Small Interactions & 50 & $528.5 \pm 59.2$ & 515.5 & $447 -- 640$ \\
         & Large Interactions & 50 & $1209.9 \pm 460.7$ & 1055.5 & $828 -- 3330$ \\
        \midrule
        \multirow{2}{*}{Review Writing} & Small Interactions & 50 & $315.7 \pm 31.0$ & 309.0 & $268 -- 372$ \\
         & Large Interactions & 50 & $558.3 \pm 173.8$ & 515.5 & $379 -- 1267$ \\
        \bottomrule
    \end{tabular*}
\end{table*}

\section{Datasets}
We conduct our experiments on LongLaMP \citep{kumar2024longlamp} benchmark. 
In this section, we provide a detailed description of the datasets used in our experiments, the preprocessing transformations applied to align them with our experimental setup, and their corresponding descriptive statistics. 

\subsection{Dataset specification}
\paragraph{Abstract Generation.}
This task assesses the ability of a model to generate scientific abstracts given paper titles and keywords. 
The user profile consists of the researcher's previous publications, which the model must leverage to personalize the academic tone, logical structure, and domain-specific vocabulary unique to each author.
We follow the \textit{User Setting} in LongLaMP, where each sample is structured such that both the profile and the target output belong to the same author.
To adapt this to our setup, we first sort each researcher's publication history chronologically based on publication year.
The history is then partitioned into five sequential periods.
Within each period, we use the earliest 90\% of interactions as observed history and hold out the most recent 10\% for evaluation.
To ensure a sufficiently long-term interaction history for drift analysis, we filter for users whose total profile size exceeds 500 publications.

\paragraph{Review Writing.}
This task involves drafting a detailed product review from a product's name and description, alongside a summary of the user's review.
The profile includes a user's past reviews and ratings, requiring the model to follow the user's subjective perspective, rating tendencies, and unique writing styles when expressing opinions.
Consistent with the \textit{User Setting}, all historical interactions are mapped to a specific reviewer ID.
We follow a similar temporal splitting strategy, dividing the history into five chronological periods.
For this task, we choose users with a total profile size exceeding 300 reviews to focus on those with rich historical data. 

\subsection{Dataset statistics}
\label{supp:profile_size}
\paragraph{User categorization criteria.}
We categorize users based on their total number of historical interactions.
To ensure a distinct contrast between the two groups, we utilize the 50th and 80th percentiles of the profile size distribution within each task as thresholds: 
\begin{itemize}[leftmargin=3.5mm, topsep=3pt, itemsep=3pt, parsep=0pt]
    \item[$\circ$] \textit{Small Interactions:} Users within the bottom 50\% of the interaction-frequency distribution, representing users with relatively shorter interaction histories.
    \item[$\circ$] \textit{Large Interactions:} Users within the top 20\% of the interaction-frequency distribution, representing users with substantially longer interaction histories.
\end{itemize}
We randomly sampled 50 users from each category for our main experiments to maintain computational efficiency.

\paragraph{Descriptive statistics.}
Table~\ref{tab:interaction_stats} presents the profile size statistics for each task and group.
The filtering thresholds above are applied to each user's full interaction history before the within-period held-out split, whereas the table reports the number of observed historical interactions after the split.
In Abstract Generation, where only users with more than 500 interactions in their full histories are included, the Large group exhibits a clearly higher mean profile size, testing the model's ability to handle substantially longer interaction histories.
In Review Writing, the Small group reflects relatively shorter interaction histories, whereas the Large group provides substantially richer histories and poses a greater challenge in filtering redundant or noisy interactions.

\section{Baselines}
In this section, we provide the implementation details and configurations for the baseline models.
To distinguish between the methods, we denote the base model with $\square$, personalization baselines with $\boxplus$, and continual learning strategies with $\blacksquare$.
Notably, as the original CL baselines were primarily designed for task-incremental or class-incremental settings, we have adapted their core mechanisms to suit our user-specific, period-wise adaptation setup.

\begin{itemize}
    \item[$\square$] \textbf{Base} denotes the original pre-trained LLM without any additional context or fine-tuning, serving as a zero-shot performance baseline.

    \item[$\boxplus$] \textbf{RAG (Retrieval-Augmented Generation)} \citep{hwang2023aligning} is a strategy that retrieves the top-$k$ relevant records from the user's history as in-context examples. In our setup, we utilize the \texttt{BM25Okapi} implementation from the \texttt{rank\_bm25} library to retrieve the top-4 interactions. We consider two variations: \textit{RAG each}, which restricts the retrieval pool to the history within the same period $t$ as the test query, and \textit{RAG accum}, which performs retrieval from the cumulative history from periods $0$ to $t$.

    \item[$\boxplus$] \textbf{PAG (Profile-Augmented Generation)} \citep{richardson2023integrating} is a method that utilizes a generated user profile as context, following the profiling strategy of prior work \citep{tan2024democratizing}. For each user, we construct a profile by randomly sampling $k=10$ historical interactions and summarizing them using \texttt{Gemma-3-4B-IT}. Similar to RAG, we evaluate two settings: \textit{PAG each}, where the 10 samples are drawn exclusively from the history within the current period $t$, and \textit{PAG accum}, where the 10 samples are selected uniformly from the cumulative history across periods $0$ to $t$.
    
    \item[$\boxplus$] \textbf{OPPU (One PEFT per User)} \citep{tan2024democratizing} originally trains individual user LoRA adapters starting from a Task Adapted Model (TAM). In our implementation, we adapt this method by directly training individual LoRA adapters for each user from the base model, bypassing the intermediate TAM stage to focus on direct user-specific optimization. We evaluate two training settings: \textit{OPPU each}, which fine-tunes the adapter only on the history of the current period $t$, and \textit{OPPU accum}, which utilizes the cumulative history from periods $0$ to $t$ for training.\looseness=1

    \item[$\blacksquare$] \textbf{CAMA (Confidence-Aware Memory Augmentation)} \citep{kim2024online} is a replay-based CL strategy that filters new samples according to the model's self-confidence. At the start of each new period, the model calculates its average confidence level using samples from the previous period. For new incoming data, only those samples that exceed this average confidence threshold are selected for fine-tuning. These high-confidence samples are integrated with a subset of past interactions from the replay buffer, ensuring that the adapter focuses on reliable information while maintaining previously established user patterns. 
    
    \item[$\blacksquare$] \textbf{SAR (Similarity-Aware Replay)} \citep{seo2024budgeted} optimizes the replay process by maintaining a balanced representation of the user's behavioral space. This method clusters user interactions in the hidden representation space and tracks a use-frequency score for each sample. To prevent the model from overfitting to repetitive patterns, SAR assigns higher sampling probabilities to interactions that are from under-represented clusters or have been less frequently used in training. By accounting for similarity and usage history, it ensures that the replay buffer provides a diverse and comprehensive coverage of the user's history.
    
    \item[$\blacksquare$] \textbf{O-LoRA (Orthogonal LoRA)} \citep{wang2023orthogonal} mitigates catastrophic forgetting by constraining learning updates for different periods to mutually orthogonal subspaces. When training on a new period, the model projects the weight updates onto the orthogonal complement of the space utilized in previous stages. This restriction prevents new knowledge from distorting the previously learned weights, allowing the model to accumulate information across multiple periods without degrading performance on earlier tasks. 
    
    \item[$\blacksquare$] \textbf{EWC-LoRA (Elastic Weight Consolidation for LoRA)} \citep{xiang2023language} is a method that preserves past knowledge by penalizing changes to parameters that were critical for previous learning stages. After each period, the model estimates the importance of each LoRA parameter based on its contribution to the past data distribution. During the training of a new period, a regularization penalty term is added to the loss function, which restricts the movement of these important parameters. This elastic constraint forces the model to learn new information primarily through less critical parameters, thereby consolidating user-specific knowledge acquired in earlier stages. 
    
\end{itemize}

\section{Additional Evaluation and Statistical Validation}
\label{sec:additional_evaluation}

We conduct additional evaluations to assess the robustness and statistical reliability of the main results beyond the point estimates reported in the main experiments.

\subsection{LLM-as-a-Judge evaluation}
\begin{table}[t]
\centering
\caption{\textbf{LLM-as-a-Judge evaluation.}
Pairwise preference results of \name{} against Base and CAMA across tasks.
Abs and Rev denote Abstract Generation and Review Writing, respectively.}
\label{tab:llm_judge}
\resizebox{\columnwidth}{!}{%
\begin{tabular}{llccc}
\toprule[1.5pt]
\textbf{Task} & \textbf{Baseline} & \textbf{Avg. Pref. Score} $\uparrow$ & \textbf{Consistent Win Rate} $\uparrow$ & \textbf{$p$-value} \\
\midrule[1pt]
\multirow{2}{*}{Abs}
& Base & 66.7\% & 88.8\% & $<10^{-10}$ \\
& CAMA    & 56.1\% & 63.7\% & $<10^{-5}$ \\
\midrule
\multirow{2}{*}{Rev}
& Base & 59.5\% & 71.7\% & $<10^{-3}$ \\
& CAMA    & 67.6\% & 83.9\% & $<10^{-10}$ \\
\bottomrule[1.5pt]
\end{tabular}%
}
\end{table}

To complement the ROUGE-based evaluation, we conduct an additional LLM-as-a-Judge evaluation using \texttt{gpt-4o-mini}~\cite{achiam2023gpt}.
We use a representative subset consisting of 20\% of users from each interaction-size group for both Abstract Generation and Review Writing, while preserving the same temporal split protocol.
We compare \name{} against Base and CAMA using pairwise preference evaluation with randomized response order to reduce position bias.
The average preference score assigns 1, 0.5, and 0 to wins, ties, and losses, respectively.
The consistent win rate is computed over order-consistent decisive comparisons, excluding ties and inconsistent judgments.
As shown in Table~\ref{tab:llm_judge}, \name{} is consistently preferred over both Base and CAMA across tasks, with consistent win rates ranging from 63.7\% to 88.8\%.
These results provide complementary evidence that the improvements of \name{} extend beyond lexical overlap measured by ROUGE.

\subsection{Multi-seed training robustness}
\begin{table}[t]
    \centering
    \small
    \caption{\textbf{Multi-seed training results on Abstract Generation.}
    Mean and standard deviation across three training seeds.
    Results are reported on the 20-user Abstract Generation subset.}
    \label{tab:multiseed}

    \begin{tabular}{l|cc}
        \toprule[1.5pt]

        \textbf{Method}
        & \textbf{ROUGE-1}
        & \textbf{ROUGE-L} \\

        \midrule[1pt]

        Base      & 0.300 & 0.146 \\
        RAG each  & 0.302 & 0.159 \\
        OPPU each & $0.324 \pm 0.005$ & $0.155 \pm 0.002$ \\
        CAMA      & $0.333 \pm 0.014$ & $0.160 \pm 0.007$ \\
        \rowcolor[HTML]{D9D9D9}
        \name{}   & $0.346 \pm 0.004$ & $0.188 \pm 0.002$ \\

        \bottomrule[1.5pt]
    \end{tabular}

\vspace{-0.05in}
\end{table}

Although inference is deterministic under greedy decoding, adapter training remains stochastic.
We therefore retrain the training-based methods with three random seeds and report the mean and standard deviation of their performance on Abstract Generation.
As shown in Table~\ref{tab:multiseed}, \name{} maintains the highest performance under different training seeds while exhibiting relatively small variation.
This indicates that the observed improvement is robust to stochasticity in adapter training.

\subsection{User-level and user-period variance}
\begin{table}[t]
    \centering
    \caption{\textbf{Performance variance across user-period instances.}
    Mean and standard deviation across user-period instances for both tasks.}
    \label{tab:user_period_variance}

    \resizebox{\columnwidth}{!}{%
        \begin{tabular}{l|cc|cc}
            \toprule[1.5pt]

            \multirow{2}{*}{\textbf{Method}}
            & \multicolumn{2}{c|}{\textbf{Abstract Generation}}
            & \multicolumn{2}{c}{\textbf{Review Writing}} \\

            \cmidrule(lr){2-3} \cmidrule(lr){4-5}
            & ROUGE-1 & ROUGE-L & ROUGE-1 & ROUGE-L \\

            \midrule[1pt]

            Base
            & $0.301 \pm 0.016$
            & $0.147 \pm 0.008$
            & $0.302 \pm 0.036$
            & $0.125 \pm 0.012$ \\

            RAG each
            & $0.301 \pm 0.018$
            & $0.159 \pm 0.011$
            & $0.276 \pm 0.039$
            & $0.123 \pm 0.014$ \\

            OPPU each
            & $0.319 \pm 0.024$
            & $0.153 \pm 0.012$
            & $0.307 \pm 0.037$
            & $0.125 \pm 0.011$ \\

            CAMA
            & $0.337 \pm 0.024$
            & $0.164 \pm 0.013$
            & $0.312 \pm 0.038$
            & $0.125 \pm 0.012$ \\

            \rowcolor[HTML]{D9D9D9}
            \name{}
            & $0.350 \pm 0.028$
            & $0.190 \pm 0.017$
            & $0.326 \pm 0.037$
            & $0.142 \pm 0.013$ \\

            \bottomrule[1.5pt]
        \end{tabular}%
    }
\vspace{-0.05in}
\end{table}
\begin{table}[t]
    \centering
    \caption{\textbf{Performance variance across users.}
    Mean and standard deviation across users for both tasks.}
    \label{tab:user_variance}

    \resizebox{\columnwidth}{!}{%
        \begin{tabular}{l|cc|cc}
            \toprule[1.5pt]

            \multirow{2}{*}{\textbf{Method}}
            & \multicolumn{2}{c|}{\textbf{Abstract Generation}}
            & \multicolumn{2}{c}{\textbf{Review Writing}} \\

            \cmidrule(lr){2-3} \cmidrule(lr){4-5}
            & ROUGE-1 & ROUGE-L & ROUGE-1 & ROUGE-L \\

            \midrule[1pt]

            Base
            & $0.301 \pm 0.008$
            & $0.147 \pm 0.005$
            & $0.302 \pm 0.031$
            & $0.125 \pm 0.010$ \\

            RAG each
            & $0.301 \pm 0.009$
            & $0.159 \pm 0.006$
            & $0.276 \pm 0.033$
            & $0.123 \pm 0.011$ \\

            OPPU each
            & $0.319 \pm 0.015$
            & $0.153 \pm 0.008$
            & $0.307 \pm 0.032$
            & $0.125 \pm 0.009$ \\

            CAMA
            & $0.337 \pm 0.012$
            & $0.164 \pm 0.007$
            & $0.312 \pm 0.032$
            & $0.125 \pm 0.010$ \\
            \rowcolor[HTML]{D9D9D9}
            \name{}
            & $0.350 \pm 0.017$
            & $0.190 \pm 0.013$
            & $0.326 \pm 0.030$
            & $0.142 \pm 0.010$ \\

            \bottomrule[1.5pt]
        \end{tabular}%
    }
\vspace{-0.05in}
\end{table}

We further examine performance variation across individual user-period instances and across users for both tasks.
For each method, we report the mean and standard deviation of ROUGE scores under the two aggregation levels.
Tables~\ref{tab:user_period_variance} and~\ref{tab:user_variance} show that \name{} achieves the highest mean performance across both tasks and metrics under both aggregation levels.
The improvements therefore persist across user histories rather than being driven by a small subset of users or periods.

\subsection{Paired bootstrap significance tests}
\begin{table}[t]
\centering
\caption{\textbf{Paired bootstrap significance tests against CAMA.}
Paired differences and 95\% confidence intervals across tasks and metrics ($N=400$).
Abs and Rev denote Abstract Generation and Review Writing, respectively.
R-1 and R-L denote ROUGE-1 and ROUGE-L, respectively.}
\label{tab:bootstrap}
\resizebox{\columnwidth}{!}{%
\begin{tabular}{llcccc}
\toprule[1.5pt]
\textbf{Task}
& \textbf{Metric}
& \textbf{CAMA}
& \textbf{\name{}}
& \textbf{Diff.}
& \textbf{95\% CI} \\
\midrule[1pt]
\multirow{2}{*}{Abs}
& R-1 & 0.337 & 0.350 & +0.013 & $[0.011,\,0.016]$ \\
& R-L & 0.164 & 0.190 & +0.026 & $[0.025,\,0.028]$ \\
\midrule
\multirow{2}{*}{Rev}
& R-1 & 0.312 & 0.326 & +0.014 & $[0.011,\,0.017]$ \\
& R-L & 0.125 & 0.142 & +0.017 & $[0.016,\,0.018]$ \\
\bottomrule[1.5pt]
\end{tabular}%
}
\end{table}

Finally, we assess the statistical significance of the improvements over CAMA, the strongest continual learning baseline, using paired bootstrap resampling with 10,000 resamples.
We repeatedly resample the same evaluation instances and compute the difference between \name{} and CAMA for each task and metric.
As shown in Table~\ref{tab:bootstrap}, the 95\% confidence intervals of the paired differences are strictly above zero for both ROUGE-1 and ROUGE-L on both tasks.
These results indicate that the improvements over CAMA are statistically stable across evaluation instances.

\section{Training-side Analyses}
\label{sec:training_analyses}

\subsection{Selection ratio.}
\begin{table}[t]
    \centering
    \caption{\textbf{Sensitivity to the selection ratio.}
    Generation performance under different selection ratios on Abstract Generation.
    Small and Large Interactions denote subsets of 10 users sampled from each interaction-size group, covering 20\% of the full evaluation users.}
    \label{tab:selection_ratio}

    \resizebox{\columnwidth}{!}{%
        \begin{tabular}{l|cc|cc}
            \toprule[1.5pt]

            \multirow{2}{*}{\textbf{Selection}}
            & \multicolumn{2}{c|}{\textbf{Small Interactions}}
            & \multicolumn{2}{c}{\textbf{Large Interactions}} \\

            \cmidrule(lr){2-3} \cmidrule(lr){4-5}
            & ROUGE-1 & ROUGE-L & ROUGE-1 & ROUGE-L \\

            \midrule[1pt]

            Top-10\%        & 0.322 & 0.152 & 0.326 & 0.157 \\
            Top-20\%        & 0.324 & 0.151 & 0.320 & 0.154 \\
            \rowcolor[HTML]{D9D9D9}
            Top-30\% (Ours) & 0.326 & 0.153 & 0.323 & 0.155 \\
            Top-50\%        & 0.329 & 0.155 & 0.312 & 0.150 \\
            All-100\%       & 0.305 & 0.147 & 0.297 & 0.147 \\

            \bottomrule[1.5pt]
        \end{tabular}%
    }
\vspace{-0.05in}
\end{table}

We further examine the proportion of interactions selected for adapter updates.
As shown in Table~\ref{tab:selection_ratio}, the optimal ratio varies with profile size.
The Small Interactions group benefits from retaining more of its limited history, whereas the Large Interactions group benefits from stronger filtering.
Top-30\% provides a robust intermediate operating point across both interaction regimes, while using all interactions consistently degrades performance.

\subsection{Drift score analysis}

\paragraph{Structural analysis and robustness to $\alpha$.}

We analyze the relationship between the \textit{Novelty Term} ($N$) and the \textit{Quality Term} ($Q$) in our \textit{Drift Score} ($D$) from Eq.~\ref{eq:drift_score}:
\begin{equation}
\label{eq:component_def}
D =
\underbrace{- \log \frac{p_{\mathrm{adap}}}{p_{\mathrm{base}}}}_{N}
+
\underbrace{\alpha \log p_{\mathrm{base}}}_{Q}.
\end{equation}

\begin{figure}[t] 
\centering
\includegraphics[width=\columnwidth]{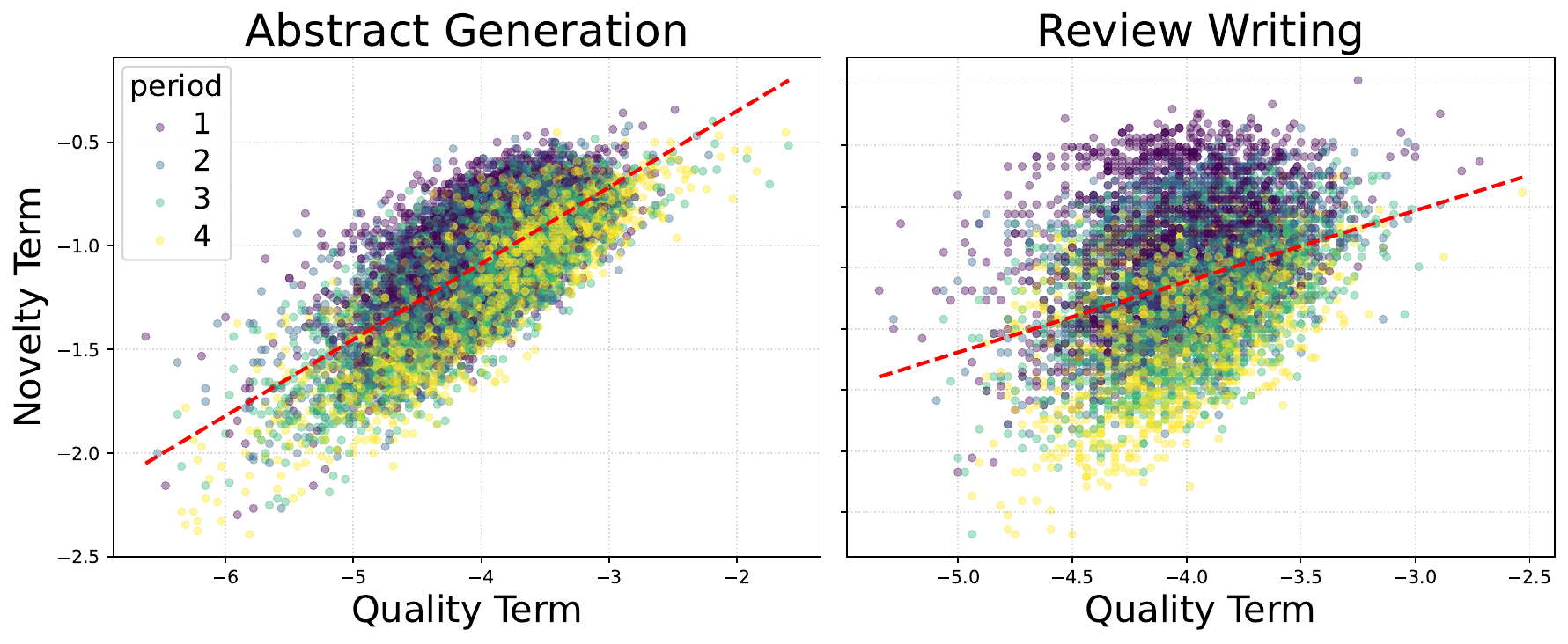}
\caption{\textbf{Correlation between quality and novelty terms.} 
The scatter plots show the relationship between the Quality Term and the Novelty Term.
Spearman’s $\rho$ is $0.6647$ for Abstract Generation and $0.3204$ for Review Writing.}
\label{fig:quality_novelty}
\end{figure}

This analysis is conducted on a representative subset consisting of 10 users from each interaction group for both tasks.
Figure~\ref{fig:quality_novelty} shows a positive correlation between $Q$ and $N$ across periods ($r=0.66$ for Abstract Generation and $r=0.32$ for Review Writing).
The relatively sparse low-quality, high-novelty region suggests that the benchmark contains few cases where novelty is primarily driven by noisy or poorly formed text.

\begin{table}[t]
    \centering
    \caption{\textbf{Robustness to the balancing parameter $\alpha$. } Performance comparison using $\alpha$ for the drift score.} 
    \label{tab:oppu_alpha}
    
    \resizebox{\columnwidth}{!}{%
        \begin{tabular}{l|cc|cc} 
            \toprule[1.5pt]
            
            \multirow{2}{*}[-2pt]{\shortstack[l]{\textbf{Balancing}\\[1pt]\textbf{Parameter}}}
            & \multicolumn{2}{c|}{\textbf{Abstract Generation}} & \multicolumn{2}{c}{\textbf{Review Writing}} \\ 
            
            \cmidrule(lr){2-3} \cmidrule(lr){4-5}
             & ROUGE-1 & ROUGE-L & ROUGE-1 & ROUGE-L \\ 
            
            \midrule[1pt]
            
            $\alpha=0.0$ & 0.324 & 0.154 & 0.314 & 0.125 \\
            $\alpha=0.5$ & 0.323 & 0.155 & 0.313 & 0.125 \\
            \rowcolor[HTML]{D9D9D9}
            $\alpha=1.0$ (Ours) & 0.324 & 0.155 & 0.313 & 0.125 \\ 
            $\alpha=\infty$ & 0.323 & 0.155 & 0.313 & 0.125 \\ 
            
            \bottomrule[1.5pt]
        \end{tabular}%
    }
\end{table}

This structural relationship also helps explain the robustness to the balancing weight $\alpha$.
As shown in Table~\ref{tab:oppu_alpha}, varying $\alpha$ from 0.0 to $\infty$ yields similar performance across tasks.
We use $\alpha=1.0$ as the default to retain an explicit quality signal when identifying drift samples.

\paragraph{Distinguishing drift score from hard example mining.}

A potential concern is whether the drift score simply performs hard example mining by targeting samples with low $p_{\mathrm{adap}}$.
We define the \textit{Hardness Term} ($H$) as
\begin{equation}
H = - \log p_{\mathrm{adap}}.
\end{equation}
Using Eq.~\ref{eq:component_def},
\begin{equation}
\label{eq:hardness_decomp_general}
H = N - \log p_{\mathrm{base}},
\end{equation}
and for $\alpha \neq 0$,
\begin{equation}
\label{eq:hardness_decomp_alpha}
H = N - \frac{Q}{\alpha}.
\end{equation}

Thus, a high hardness score can arise either from under-adaptation or from low-quality instances that are difficult for both the base and adapted models.
Hardness alone therefore does not isolate the mismatch that our drift score is designed to capture.

\begin{figure}[t] 
\centering
\includegraphics[width=\columnwidth]{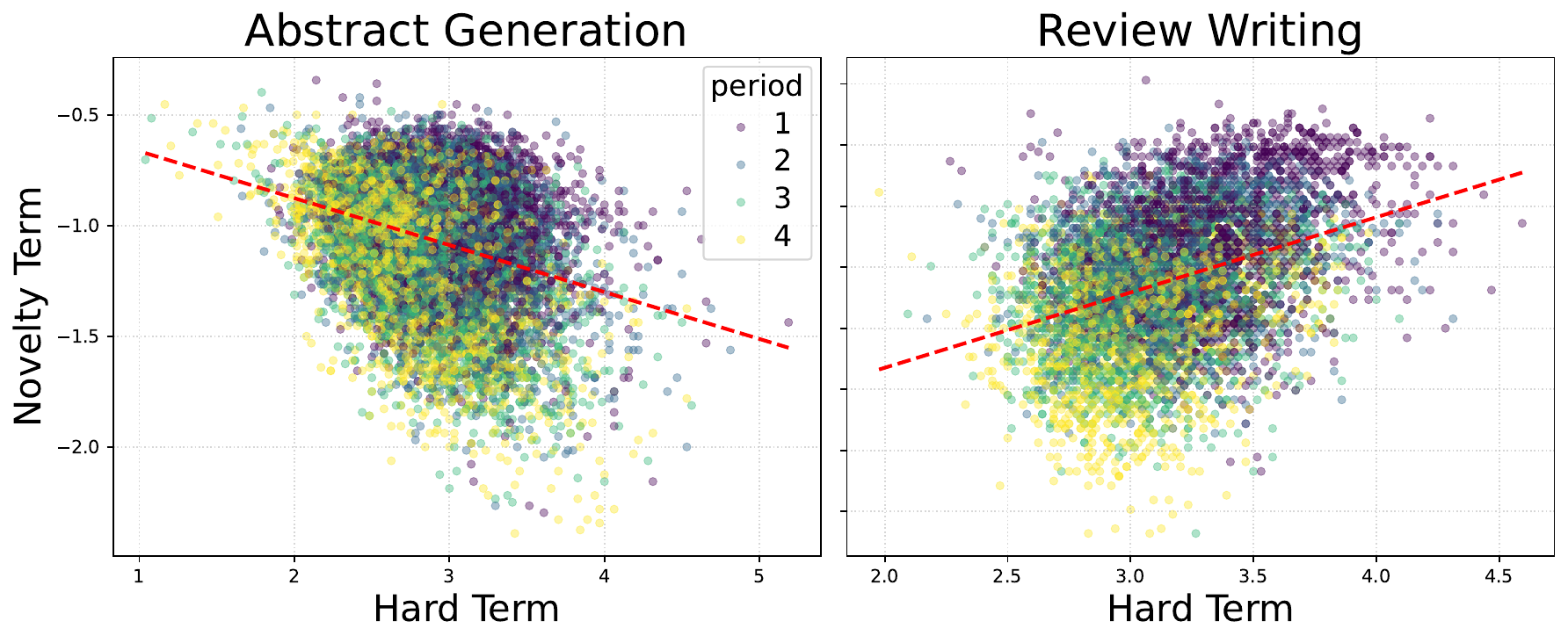}
\caption{\textbf{Correlation between hardness and novelty terms.} 
The scatter plots show the relationship between the Hardness Term and the Novelty Term.
Spearman’s $\rho$ is $-0.2361$ for Abstract Generation and $0.3468$ for Review Writing.}
\label{fig:hard_novelty}
\end{figure}

\begin{figure}[t] 
\centering
\includegraphics[width=\columnwidth]{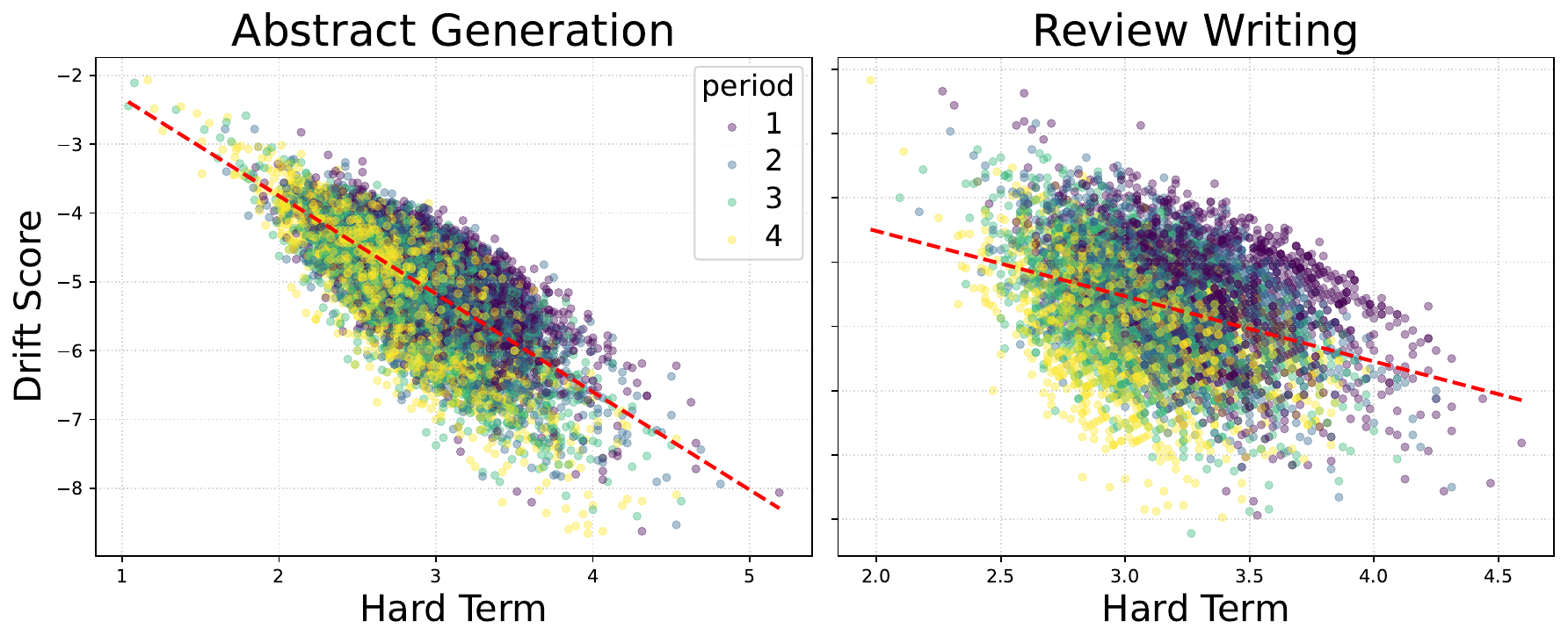}
\caption{\textbf{Correlation between hardness term and drift score.} 
The scatter plots show the relationship between the Hardness Term and the Drift Score. 
Spearman’s $\rho$ is $-0.6997$ for Abstract Generation and $-0.3421$ for Review Writing.}
\label{fig:hard_drift}
\vspace{-0.05in}
\end{figure}

\begin{figure}[t] 
\centering
\includegraphics[width=\columnwidth]{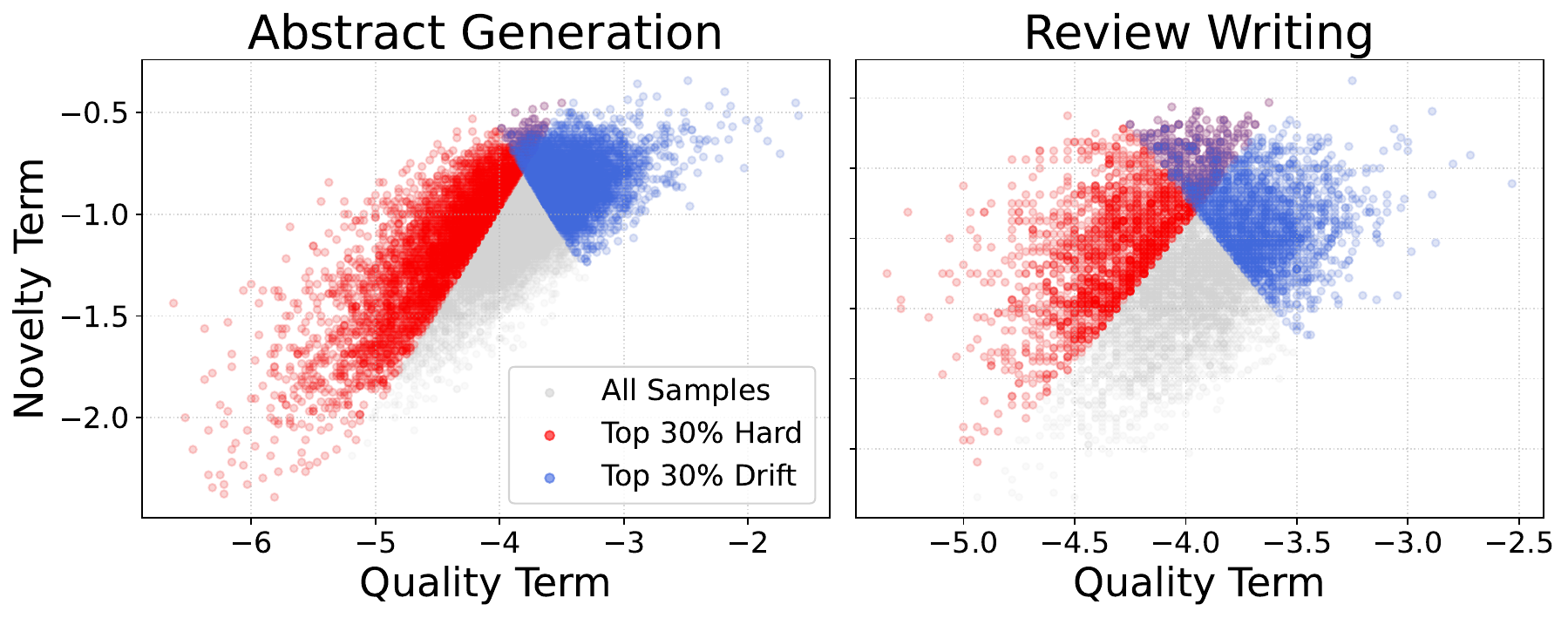}
\caption{\textbf{Visualization of sample selections by hardness term versus drift score.}
Gray points denote all samples, while red and blue highlight the top 30\% samples by the Hardness Term and the Drift Score, respectively.}
\label{fig:visual_scores}
\end{figure}

Empirically, as shown in Figures~\ref{fig:hard_novelty} and~\ref{fig:hard_drift}, the relationship between $H$ and $N$ differs across tasks, while $H$ and $D$ are negatively correlated in both tasks.
Figure~\ref{fig:visual_scores} further shows that hardness-based selection extends into lower-quality regions, whereas drift-based selection is concentrated among samples with both stronger quality and novelty signals.
Together, these results distinguish the proposed drift score from conventional hard example mining.

\section{Inference-side Analyses}
\label{sec:inference_analyses}

\subsection{Replay buffer retention policies}
\label{sec:appendix_buffer}

We examine four replay buffer retention policies under a fixed budget $N_{\max}$.
\textit{Global Highest} retains samples with the highest scores, \textit{Cluster-wise} prioritizes clusters with high average scores, \textit{Cluster Round} selects samples across clusters in a round-robin manner, and \textit{Random} uniformly samples from the candidate pool.
For cluster-based strategies, we use Sentence-BERT \citep{reimers2019sentence} with \texttt{all-mpnet-base-v2} and determine the number of $K$-Means clusters by the Silhouette Score over $K \in [2,10]$.

\begin{table}[t]
    \centering
    \caption{\textbf{Effectiveness of replay buffer retention policies.}
    Comparison of retention strategies on the 20-user subsets of Abstract Generation and Review Writing.}
    \label{tab:buffer}

    \resizebox{\columnwidth}{!}{%
        \begin{tabular}{l|cc|cc}
            \toprule[1.5pt]

            \multirow{3}{*}[-2pt]{\shortstack[l]{\textbf{Buffer Retention}\\[1pt]\textbf{Policy}}}
            & \multicolumn{4}{c}{\textbf{Abstract Generation}} \\
            \cmidrule(lr){2-5}
            & \multicolumn{2}{c|}{ROUGE-1}
            & \multicolumn{2}{c}{ROUGE-L} \\
            \cmidrule(lr){2-3} \cmidrule(lr){4-5}
            & $\sigma=0$ & $\sigma=\text{Q3}$
            & $\sigma=0$ & $\sigma=\text{Q3}$ \\

            \midrule[1pt]
            \rowcolor[HTML]{D9D9D9}
            Global Highest (Ours) & 0.351 & 0.350 & 0.191 & 0.189 \\
            Cluster Round         & 0.332 & 0.346 & 0.185 & 0.188 \\
            Cluster-wise          & 0.333 & 0.345 & 0.186 & 0.188 \\
            Random                & 0.335 & 0.345 & 0.187 & 0.187 \\

            \midrule[1.5pt]

            \multirow{3}{*}[-2pt]{\shortstack[l]{\textbf{Buffer Retention}\\[1pt]\textbf{Policy}}}
            & \multicolumn{4}{c}{\textbf{Review Writing}} \\
            \cmidrule(lr){2-5}
            & \multicolumn{2}{c|}{ROUGE-1}
            & \multicolumn{2}{c}{ROUGE-L} \\
            \cmidrule(lr){2-3} \cmidrule(lr){4-5}
            & $\sigma=0$ & $\sigma=\text{Q3}$
            & $\sigma=0$ & $\sigma=\text{Q3}$ \\

            \midrule[1pt]
            \rowcolor[HTML]{D9D9D9}
            Global Highest (Ours) & 0.322 & 0.332 & 0.144 & 0.145 \\
            Cluster Round         & 0.322 & 0.331 & 0.145 & 0.145 \\
            Cluster-wise          & 0.323 & 0.332 & 0.145 & 0.145 \\
            Random                & 0.324 & 0.331 & 0.145 & 0.145 \\

            \bottomrule[1.5pt]
        \end{tabular}%
    }

\vspace{-0.05in}
\end{table}

Table~\ref{tab:buffer} compares these policies under $\sigma=0$ and our relevance-gated setting.
\textit{Global Highest} performs best on Abstract Generation and remains competitive on Review Writing, where differences among retention strategies are relatively small.
These results support score-based retention as a robust default while indicating limited sensitivity to the exact retention policy.

\subsection{Interpolation coefficients}

\begin{table}[t]
    \centering
    \caption{\textbf{Effectiveness of interpolation.} Performance comparison across different weighting factors ($\lambda$). The coefficient $\lambda$ determines the relative reliance on retrieved context versus the model's internal parametric knowledge.} 
    \label{tab:inf_ablation1}
    
    \resizebox{\columnwidth}{!}{%
        \begin{tabular}{l|cc|cc} 
            \toprule[1.5pt]
            
            \multirow{2}{*}[-2pt]{\shortstack[l]{\textbf{Interp.}\\[1pt]\textbf{Coefficient}}}
            & \multicolumn{2}{c|}{\textbf{Abstract Generation}} & \multicolumn{2}{c}{\textbf{Review Writing}} \\ 
            
            \cmidrule(lr){2-3} \cmidrule(lr){4-5}
             & ROUGE-1 & ROUGE-L & ROUGE-1 & ROUGE-L \\ 
            
            \midrule[1pt]
            
            $\lambda=0.0$ & 0.324 & 0.155 & 0.313 & 0.125 \\
            $\lambda=0.1$ & 0.334 & 0.187 & 0.326 & 0.142 \\
            \rowcolor[HTML]{D9D9D9}
            $\lambda=0.5$ (Ours) & 0.350 & 0.190 & 0.326 & 0.142 \\ 
            $\lambda=1.0$ & 0.294 & 0.154 & 0.276 & 0.112 \\
            
            \bottomrule[1.5pt]
        \end{tabular}%
    }
    \vspace{-0.1in}
\end{table}

Table~\ref{tab:inf_ablation1} examines the interpolation coefficient $\lambda$ from Eq.~\ref{eq:interp}.
Relying exclusively on either the parametric path ($\lambda=0.0$) or the retrieved context ($\lambda=1.0$) is suboptimal across tasks.
Moderate interpolation performs best, with $\lambda=0.5$ improving ROUGE-L over both single-path settings.
These results support $\lambda=0.5$ as a default for combining parametric adaptation with retrieved user history.

\section{Component Scaling}
\label{sec:component_scaling}

\subsection{Replay buffer size}
\begin{table}[t]
    \centering
    \small
    \caption{\textbf{Sensitivity to the replay buffer size.}
    Performance comparison across different replay buffer budgets. Results are reported on the 20-user Abstract Generation subset.}
    \label{tab:scaling_buffer}

    \begin{tabular}{l|cc}
        \toprule[1.5pt]
        \textbf{Replay Buffer Size}
        & \textbf{ROUGE-1}
        & \textbf{ROUGE-L} \\
        \midrule[1pt]
        $N_{\max}=25$        & 0.338 & 0.184 \\
        \rowcolor[HTML]{D9D9D9}
        $N_{\max}=50$ (Ours) & 0.350 & 0.189 \\
        $N_{\max}=100$       & 0.346 & 0.187 \\
        \bottomrule[1.5pt]
    \end{tabular}

\vspace{-0.05in}
\end{table}

We examine the sensitivity of \name{} to the replay buffer budget $N_{\max}$.
As shown in Table~\ref{tab:scaling_buffer}, increasing the buffer size from 25 to 50 improves both ROUGE-1 and ROUGE-L, while increasing it further to 100 provides no additional benefit.
This supports $N_{\max}=50$ as a cost-effective default.

\subsection{Retrieved history size}
\begin{table}[t]
    \centering
    \caption{\textbf{Sensitivity to the retrieved history size.} Performance comparison across different numbers of retrieved interactions ($k$). Results are reported on the 20-user Abstract Generation subset.}
    \label{tab:scaling_k}
    
    \resizebox{0.85\columnwidth}{!}{%
        \begin{tabular}{l|cc} 
            \toprule[1.5pt]
            \textbf{Retrieved Interaction Size} & \textbf{ROUGE-1} & \textbf{ROUGE-L} \\
            \midrule[1pt]
            $k=1$ & 0.326 & 0.182 \\
            \rowcolor[HTML]{D9D9D9}
            $k=2$ (Ours) & 0.350 & 0.189 \\
            $k=4$ & 0.350 & 0.192 \\
            \bottomrule[1.5pt]
        \end{tabular}%
    }
    \vspace{-0.1in}
\end{table}

Table~\ref{tab:scaling_k} examines the number of retrieved interactions $k$.
Increasing $k$ from 1 to 2 substantially improves performance, while increasing it further to 4 provides only limited additional gain.
We therefore use $k=2$ as a practical operating point that balances personalization quality and inference cost.

\subsection{Adapter rank size}
\begin{table}[t]
    \centering
\caption{\textbf{Sensitivity to the adapter rank.}
Performance comparison under different adapter ranks $r$. Results are reported on the 20-user Abstract Generation subset.}
\label{tab:scaling_r}
    
    \resizebox{0.85\columnwidth}{!}{%
        \begin{tabular}{l|cc}
            \toprule[1.5pt]
            \textbf{Method} & \textbf{ROUGE-1} & \textbf{ROUGE-L} \\
            \midrule[1pt]
            \rowcolor[HTML]{D9D9D9}
            OPPU each ($r{=}8$)  & 0.318 & 0.153 \\
            OPPU each ($r{=}16$)       & 0.245 & 0.131 \\
            OPPU each ($r{=}32$)       & 0.244 & 0.132 \\
            \midrule
            \rowcolor[HTML]{D9D9D9}
            \name{} ($r{=}8$, ours)    & 0.350 & 0.189 \\
            \name{} ($r{=}16$)         & 0.332 & 0.178 \\
            \name{} ($r{=}32$)         & 0.336 & 0.181 \\
            \bottomrule[1.5pt]
        \end{tabular}%
    }
\end{table}

Table~\ref{tab:scaling_r} examines the LoRA rank $r$.
Performance is highest at $r=8$ for both OPPU and \name{}, while larger ranks lead to lower performance.
These results indicate that a small adapter capacity is sufficient in our setting and support $r=8$ as the default configuration.


\section{Privacy and Memorization Risk Analyses}
\label{sec:privacy_memorization}

\paragraph{Evaluation protocol for memorization checks.}
For each user and period, we partition the user-specific interactions into two disjoint subsets.
We define \textit{Adapter-train} as the subset used for continual adapter updates, and \textit{Held-out} as a disjoint subset that is never used for adapter training in any period.
We then evaluate generation quality separately on queries from \textit{Adapter-train} and \textit{Held-out}.
A consistently higher score on \textit{Adapter-train} than on \textit{Held-out}, especially if the gap grows over periods, would be consistent with persistent memorization rather than generalized adaptation.

\subsection{Memorization risk under adapter-only generation}

\begin{table}[t]
    \centering
    \caption{\textbf{Gap analysis for memorization risk in the adapter-only setting.} Performance comparison between \textit{Adapter-train} and a disjoint \textit{Held-out} subset across periods. Results are reported on the 20-user Abstract Generation subset. R-1 and R-L represent ROUGE-1 and ROUGE-L, respectively.}

    \label{tab:privacy_memorization_gap}
    
    \resizebox{\columnwidth}{!}{%
    \begin{tabular}{l|cc|cc|cc|cc|cc}
        \toprule[1.5pt]
        \multirow{2}{*}{\textbf{Split}} 
        & \multicolumn{2}{c|}{\textbf{Period 1}} 
        & \multicolumn{2}{c|}{\textbf{Period 2}} 
        & \multicolumn{2}{c|}{\textbf{Period 3}} 
        & \multicolumn{2}{c|}{\textbf{Period 4}} 
        & \multicolumn{2}{c}{\textbf{Period Avg.}} \\
        \cmidrule(lr){2-3} \cmidrule(lr){4-5} \cmidrule(lr){6-7} \cmidrule(lr){8-9} \cmidrule(lr){10-11}
        & R-1 & R-L & R-1 & R-L & R-1 & R-L & R-1 & R-L & R-1 & R-L \\
        \midrule[1pt]
        Adapter-train & 0.257 & 0.141 & 0.257 & 0.134 & 0.262 & 0.132 & 0.268 & 0.135 & 0.261 & 0.135 \\
        Held-out & 0.257 & 0.136 & 0.258 & 0.134 & 0.264 & 0.134 & 0.265 & 0.134 & 0.261 & 0.135 \\
        \bottomrule[1.5pt]
    \end{tabular}%
    }
\vspace{-0.1in}
\end{table}

Under the \textit{Adapter-only} setting, we keep the base model frozen and update only the continual adapter using the selected 30\% subset per period.
We then compare performance on \textit{Adapter-train} versus \textit{Held-out} as described above.
If personalization were primarily driven by persistent memorization, we would expect \textit{Adapter-train} to consistently outperform \textit{Held-out}, potentially with a widening gap over periods.
However, Table~\ref{tab:privacy_memorization_gap} shows no systematic advantage for \textit{Adapter-train}, with nearly identical average performance across the two splits.
This suggests that the observed gains are better explained by generalized adaptation to user-specific traits than by memorization of individual training instances.

\subsection{Memorization risk under retrieval-interpolated generation}

\begin{table}[t]
\centering
\caption{\textbf{Gap analysis for memorization risk in the adapter + interpolation setting.} Performance comparison between \textit{Adapter-train} and a disjoint \textit{Held-out} subset in the \textit{Adapter + Interp.} setting across periods. Results are reported on the 20-user Abstract Generation subset.}
\label{tab:privacy_replay_memorization_gap}

\resizebox{\columnwidth}{!}{%
\begin{tabular}{l|cc|cc|cc|cc|cc}
    \toprule[1.5pt]
    \multirow{2}{*}{\textbf{Split}} 
    & \multicolumn{2}{c|}{\textbf{Period 1}} 
    & \multicolumn{2}{c|}{\textbf{Period 2}} 
    & \multicolumn{2}{c|}{\textbf{Period 3}} 
    & \multicolumn{2}{c|}{\textbf{Period 4}} 
    & \multicolumn{2}{c}{\textbf{Period Avg.}} \\
    \cmidrule(lr){2-3} \cmidrule(lr){4-5} \cmidrule(lr){6-7} \cmidrule(lr){8-9} \cmidrule(lr){10-11}
    & R-1 & R-L & R-1 & R-L & R-1 & R-L & R-1 & R-L & R-1 & R-L \\
    \midrule[1pt]
    Adapter-train & 0.363 & 0.205 & 0.356 & 0.201 & 0.342 & 0.195 & 0.355 & 0.201 & 0.354 & 0.200 \\
    Held-out      & 0.368 & 0.206 & 0.350 & 0.198 & 0.342 & 0.192 & 0.343 & 0.193 & 0.351 & 0.197 \\
    \bottomrule[1.5pt]
\end{tabular}%
}
\end{table}

We further examine whether retrieval-interpolated generation using buffered samples introduces additional memorization risk beyond adapters alone.
In our framework, buffered samples are used only as retrieved context at inference time and do not contribute additional gradient updates.
The base model remains frozen, and the adapter is updated only on the selected training subset.
Thus, retrieval does not directly introduce additional weight-based memorization.

We empirically validate this by repeating the same split-based comparison under the \textit{Adapter + Interp.} setting in Table~\ref{tab:privacy_replay_memorization_gap}.
If retrieval-interpolated generation amplified persistent memorization, we would expect a consistent advantage for \textit{Adapter-train}, potentially increasing over periods.
Instead, the relative performance varies across periods and the average gap remains small, with differences of only 0.003 in both ROUGE-1 and ROUGE-L.
These results do not show a systematic performance gap consistent with increased memorization under retrieval-interpolated generation in our current setup.

\section{Robustness under Abrupt Preference Shifts}
\label{sec:abrupt_shift}
\begin{table}[t]
    \centering
    \caption{\textbf{Robustness under abrupt preference shifts.} Performance comparison between the original trajectory (\textit{Orig}) and a swapped trajectory (\textit{Swapped}) that induces an abrupt shift and subsequent reversion. Results are reported on the 20-user Abstract Generation subset.}
    \label{tab:abrupt_shift}
    
    \resizebox{\columnwidth}{!}{%
    \begin{tabular}{l|cc|cc|cc}
        \toprule[1.5pt]
        \multirow{2}{*}{\textbf{Method}}
        & \multicolumn{2}{c|}{\textbf{Period 2}}
        & \multicolumn{2}{c|}{\textbf{Period 3}}
        & \multicolumn{2}{c}{\textbf{Period 4}} \\
        \cmidrule(lr){2-3} \cmidrule(lr){4-5} \cmidrule(lr){6-7}
        & R-1 & R-L
        & R-1 & R-L
        & R-1 & R-L \\
        \midrule[1pt]
        Base             & 0.304 & 0.149 & 0.298 & 0.147 & 0.298 & 0.144 \\
        RAG each             & 0.305 & 0.163 & 0.306 & 0.161 & 0.299 & 0.155 \\
        OPPU each             & 0.319 & 0.156 & 0.313 & 0.151 & 0.320 & 0.150 \\
        CAMA             & 0.344 & 0.170 & 0.338 & 0.165 & 0.339 & 0.162 \\
        \rowcolor[HTML]{D9D9D9}
        \name{} (Orig)    & 0.355 & 0.194 & 0.359 & 0.195 & 0.349 & 0.189 \\
        \rowcolor[HTML]{D9D9D9}
        \name{} (Swapped) & 0.350 & 0.194 & 0.352 & 0.196 & 0.346 & 0.189 \\
        \bottomrule[1.5pt]
    \end{tabular}%
}
\end{table}

To examine robustness to abrupt rather than gradual preference changes, we construct a mixed-user trajectory.
Periods 0--1 use the original data of target user A, Periods 2--3 are replaced with data from a randomly selected user B, and Period 4 returns to user A to test recovery after reversion.
We compare this swapped trajectory (\textit{\name{} (Swapped)}) with the original trajectory (\textit{\name{} (Orig)}).

Table~\ref{tab:abrupt_shift} shows that performance under the swapped trajectory remains close to the original trajectory during both the abrupt-shift periods (P2--P3) and the reversion period (P4).
ROUGE-L remains nearly unchanged, while the maximum ROUGE-1 difference is only 0.007.
Moreover, \name{} remains competitive with the baselines during the shifted periods and recovers after returning to the original user.
These results suggest that the framework remains robust under abrupt preference shifts and subsequent reversion.

\section{Evaluation on an Additional Dataset}
\label{sec:diverse_dataset}

\begin{table}[t]
    \centering
    \caption{\textbf{Evaluation on an additional Amazon review split.} Performance on three category-specific user streams constructed from the raw Amazon Product Reviews dataset. R-1 and R-L represent ROUGE-1 and ROUGE-L, respectively.}
    \label{tab:amazon_split}
    
    \resizebox{\columnwidth}{!}{%
        \begin{tabular}{l|cc|cc|cc}
            \toprule[1.5pt]
            \multirow{2}{*}{\textbf{Method}} 
            & \multicolumn{2}{c|}{\textbf{Books}} 
            & \multicolumn{2}{c|}{\textbf{CDs \& Vinyl}} 
            & \multicolumn{2}{c}{\textbf{Movies \& TV}} \\
            \cmidrule(lr){2-3} \cmidrule(lr){4-5} \cmidrule(lr){6-7}
            & R-1 & R-L & R-1 & R-L & R-1 & R-L \\
            \midrule[1pt]
            Base            & 0.297 & 0.127 & 0.298 & 0.128 & 0.285 & 0.123 \\
            RAG each        & 0.271 & 0.127 & 0.275 & 0.128 & 0.263 & 0.122 \\
            OPPU each       & 0.306 & 0.128 & 0.302 & 0.128 & 0.296 & 0.124 \\
            CAMA            & 0.290 & 0.125 & 0.290 & 0.127 & 0.280 & 0.121 \\
            \rowcolor[HTML]{D9D9D9}
            \name{}  & 0.327 & 0.145 & 0.324 & 0.145 & 0.313 & 0.141 \\
            \bottomrule[1.5pt]
        \end{tabular}%
    }
\end{table}

We additionally evaluate \name{} in a more diverse and less curated setting constructed from the raw Amazon Product Reviews dataset \citep{ni2019justifying}, beyond the LongLaMP benchmark used in our main experiments \citep{kumar2024longlamp}.
Following the category split adopted by \citet{zhao2026nextquill}, we consider three review categories: Books, CDs and Vinyl, and Movies and TV, and construct user-level interaction streams within each category.
For each category, we sample 20 users with more than 200 reviews and retain only reviews of products belonging to the corresponding category.
This produces category-consistent user histories while preserving substantial diversity in products, writing styles, and user preferences.
The resulting setting allows us to examine whether \name{} remains effective on raw and heterogeneous interaction histories beyond the more curated LongLaMP setup.

Table~\ref{tab:amazon_split} reports the results across the three categories.
\name{} achieves the best performance in both ROUGE-1 and ROUGE-L across all categories.
Compared with the strongest baseline, \name{} improves ROUGE-1 by 6.9\% on Books, 7.3\% on CDs \& Vinyl, and 5.7\% on Movies \& TV.
The corresponding ROUGE-L improvements are 13.3\%, 13.3\%, and 13.7\%, respectively.
The results also highlight the difficulty of personalization in this less curated setting.
Naive retrieval consistently underperforms the non-personalized Base model in ROUGE-1, suggesting that directly retrieved histories can contain weakly aligned or noisy interactions.
Purely parametric personalization and continual learning baselines also provide limited gains, with the continual learning baseline remaining below Base across categories.
In contrast, \name{} consistently improves over these alternatives, suggesting that selective parametric adaptation together with controlled use of retrieved history is particularly useful when user interaction streams are heterogeneous and potentially noisy.
Overall, these results demonstrate that the gains of \name{} extend beyond LongLaMP to a more challenging raw-data setting.

\section{Cost Analyses}
\label{sec:appendix_cost_analysis}

\subsection{Per-user storage}

Our framework incurs only modest per-user storage overhead because the frozen base model is shared across users.
The user-specific footprint consists only of the latest adapter snapshot and a bounded retrieval buffer.
With rank $r{=}8$, a single LoRA adapter requires 12.33
\,MB in fp32 or approximately 6.2\,MB in bf16.
Under our deployment assumption, only the latest adapter snapshot is maintained for each user.
In addition, the bounded retrieval buffer occupies approximately 0.3\,MB per user in our experiments.
Together, these results indicate that the user-specific storage cost remains lightweight and practical at scale.

\subsection{Inference overhead under relevance gating}
\label{sec:appendix_inf_overhead}

\begin{figure}[t]
\centering
\vspace{-0.1in}
\includegraphics[width=\columnwidth]{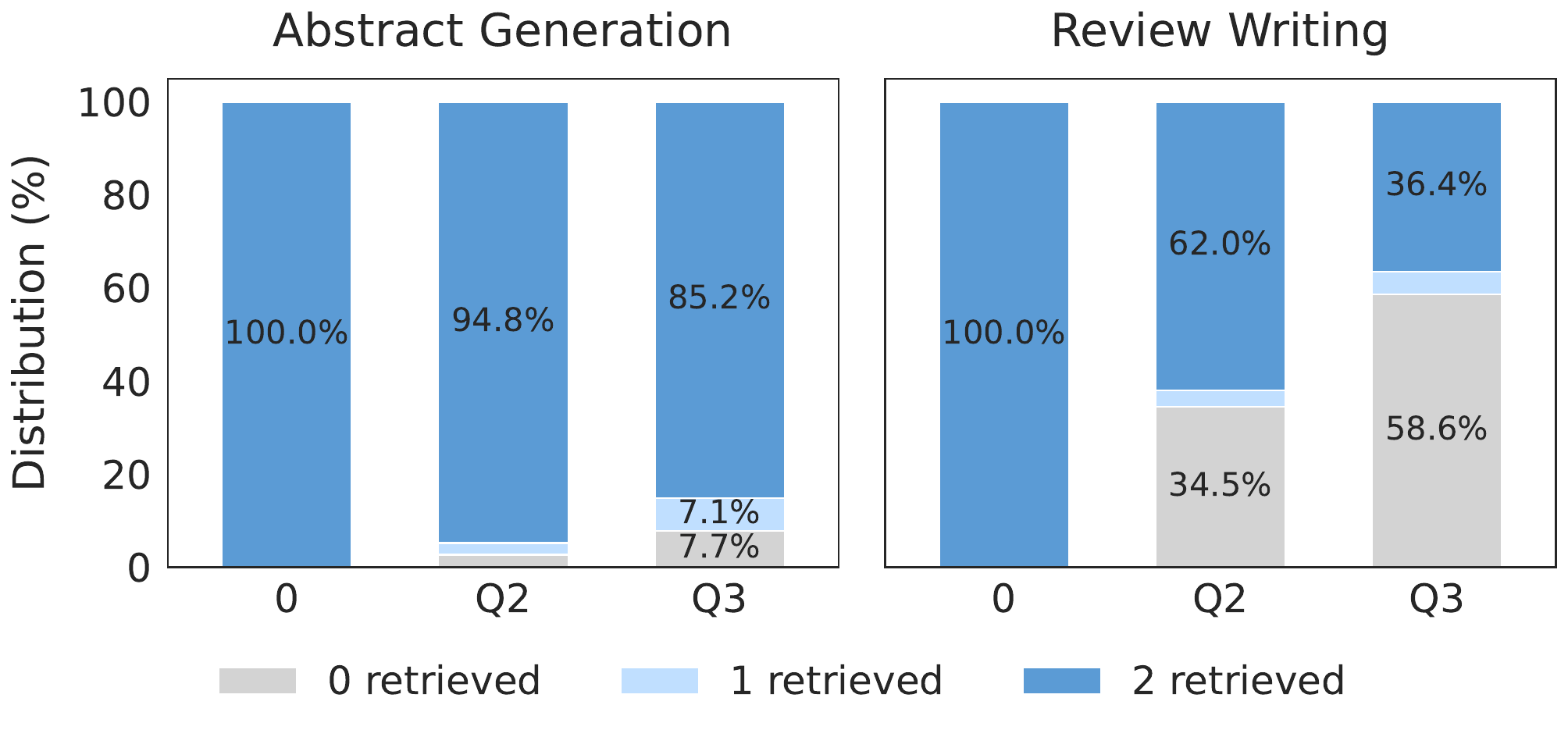}
\vspace{-0.2in}
\caption{\textbf{Distribution of retrieved interactions.} Retrieval coverage under three relevance threshold settings: no threshold ($\sigma=0$), Q2, and Q3 (ours).}
\label{fig:threshold}
\end{figure}
\begin{table}[t]
\centering
\caption{\textbf{Sensitivity to the relevance threshold.}
Performance comparison under different relevance thresholds $\sigma$ on the 20-user subsets of Abstract Generation and Review Writing.}
\label{tab:interp_ratio_threshold}

\resizebox{\columnwidth}{!}{%
\begin{tabular}{l|ccc}
\toprule[1.5pt]

\multirow{2}{*}[-2pt]{\textbf{Threshold}}
& \multicolumn{3}{c}{\textbf{Abstract Generation}} \\
\cmidrule(lr){2-4}
& Interp. (\%)
& ROUGE-1
& ROUGE-L \\

\midrule[1pt]

$\sigma{=}0.0$             & 100.0 & 0.348 & 0.189 \\
$\sigma{=}9.0$ (Q2)        & 97.2  & 0.347 & 0.189 \\
\rowcolor[HTML]{D9D9D9}
$\sigma{=}10.0$ (Q3, Ours) & 91.4  & 0.350 & 0.189 \\
$\sigma{=}10.8$            & 87.2  & 0.340 & 0.186 \\
$\sigma{=}11.8$            & 77.9  & 0.336 & 0.185 \\
$\sigma{=}12.8$            & 64.9  & 0.331 & 0.182 \\
$\sigma{=}14.0$            & 49.7  & 0.324 & 0.179 \\
$\sigma{=}15.8$            & 31.0  & 0.318 & 0.176 \\

\midrule[1.5pt]

\multirow{2}{*}[-2pt]{\textbf{Threshold}}
& \multicolumn{3}{c}{\textbf{Review Writing}} \\
\cmidrule(lr){2-4}
& Interp. (\%)
& ROUGE-1
& ROUGE-L \\

\midrule[1pt]

$\sigma{=}0.0$             & 100.0 & 0.322 & 0.144 \\
$\sigma{=}35.0$            & 93.0  & 0.317 & 0.140 \\
$\sigma{=}40.0$            & 84.8  & 0.319 & 0.140 \\
$\sigma{=}57.0$ (Q2)       & 54.5  & 0.327 & 0.145 \\
$\sigma{=}70.0$            & 51.6  & 0.323 & 0.141 \\
\rowcolor[HTML]{D9D9D9}
$\sigma{=}80.0$ (Q3, Ours) & 41.1  & 0.331 & 0.145 \\
$\sigma{=}93.0$            & 30.2  & 0.327 & 0.140 \\

\bottomrule[1.5pt]
\end{tabular}%
}

\vspace{-0.05in}
\end{table}

Dual-path interpolation increases inference cost relative to single-path decoding because activating the retrieval-conditioned path requires an additional forward pass.
However, this overhead is not incurred for every query.
The retrieval-conditioned path is executed only when the retrieved evidence passes the relevance threshold $\sigma$.
Otherwise, the model falls back to single-path generation.
The realized inference cost therefore depends on how frequently the retrieval branch is activated.

We quantify this behavior using the \textit{interpolation ratio}, defined as the percentage of queries that activate the retrieval branch and use dual-path interpolation.
Figure~\ref{fig:threshold} further illustrates how the number of retained interactions changes under different relevance thresholds.
In Abstract Generation, most queries continue to retrieve relevant interactions even under stricter thresholds.
In contrast, Review Writing exhibits a substantially larger shift toward zero retrieved interactions as the threshold increases, suggesting that historical reviews are more frequently weakly aligned with the current query.

On the 20-user analysis subset, Table~\ref{tab:interp_ratio_threshold} reports the corresponding trade-off between interpolation usage and generation quality.
For Abstract Generation, the interpolation ratio decreases from 100.0\% at $\sigma{=}0.0$ to 91.4\% at the Q3 operating point $\sigma{=}10.0$, while ROUGE-1 improves from 0.348 to 0.350 and ROUGE-L remains unchanged at 0.189.
More aggressive thresholds further reduce interpolation usage, reaching 64.9\% at $\sigma{=}12.8$ and 31.0\% at $\sigma{=}15.8$, but eventually lead to lower generation quality.

The effect of relevance gating is more substantial for Review Writing.
At the Q3 operating point $\sigma{=}80.0$, only 41.1\% of queries activate interpolation, while achieving the best ROUGE-1 score of 0.331 and matching the best ROUGE-L score of 0.145.
These results show that relevance gating can substantially reduce unnecessary retrieval-conditioned computation while preserving strong personalization performance.

\subsection{Quality--cost trade-off and optimization strategies}
\label{sec:appendix_optim}

To further reduce dual-path computation, we investigate two practical strategies that selectively limit interpolation during generation.
\begin{table}[t]
    \centering
    \caption{\textbf{Quality--cost trade-off.} Performance under early-exit and adaptive interpolation. Results are reported on the 20-user Abstract Generation subset.}
    \label{tab:optim_curve}
    
    \resizebox{\columnwidth}{!}{%
        \begin{tabular}{l|ccc}
            \toprule[1.5pt]
            \textbf{Method} & \textbf{Interp. (\%)} & \textbf{R-1} & \textbf{R-L} \\
            \midrule[1pt]
            \name{} ($\sigma=0$) & 100.0 & 0.348 & 0.189 \\
            \name{} ($\sigma=\text{Q3}$) & 91.4 & 0.350 & 0.189 \\
            Adaptive (0.9--1.0) & 73.5 & 0.337 & 0.186 \\
            Early-exit (400) & 66.8 & 0.327 & 0.182 \\
            Adaptive (0.8--0.9) & 37.0 & 0.336 & 0.186 \\
            Early-exit (200) & 33.4 & 0.330 & 0.184 \\
            Single-path (Adapter-only) & 0.0 & 0.324 & 0.154 \\
            \bottomrule[1.5pt]
        \end{tabular}%
    }
    \vspace{-0.1in}
\end{table}

\begin{itemize}[leftmargin=3.5mm, topsep=3pt, itemsep=3pt, parsep=0pt]
\item[$\circ$] \textit{Early-exit interpolation:}
Limits dual-path interpolation to the first $N$ decoding steps and then switches to single-path decoding for the remaining steps.
This provides a deterministic control over the amount of dual-path computation.
\item[$\circ$] \textit{Adaptive interpolation:}
Dynamically activates dual-path decoding according to the confidence of the single-path prediction.
Given a confidence band $(\ell,h)$, interpolation is activated when confidence falls below $\ell$ and deactivated once it exceeds $h$.
\end{itemize}
Table~\ref{tab:optim_curve} reports generation quality together with the interpolation ratio, which serves as a proxy for relative inference cost.
Relevance gating alone reduces unnecessary dual-path usage without degrading generation quality.
The additional decoding strategies provide more aggressive operating points by further limiting interpolation at the token level.

For example, Adaptive (0.8--0.9) reduces the interpolation ratio from 91.4\% to 37.0\% while retaining most of the generation quality.
As interpolation becomes less frequent, performance degrades gradually rather than immediately falling to the single-path baseline.
Moreover, the interpolation-based variants remain stronger than the Adapter-only setting, indicating that selectively applying dual-path decoding can preserve much of the benefit of retrieval-interpolated generation with substantially less computation.

\subsection{End-to-end inference cost}
\label{sec:appendix_end_to_end_cost}

\begin{table}[t]
\centering
\caption{\textbf{End-to-end inference cost.}
ROUGE scores, latency (s/token), and peak GPU memory (GB) measured on a single NVIDIA RTX 5090 with \texttt{Gemma-3-4B-IT} using a 20\% subset of Small Interaction users from each task.}
\label{tab:end_to_end_cost}

\resizebox{\columnwidth}{!}{%
\begin{tabular}{l|cccc}
\toprule[1.5pt]

\multirow{2}{*}[-2pt]{\textbf{Method}}
& \multicolumn{4}{c}{\textbf{Abstract Generation}} \\
\cmidrule(lr){2-5}
& R-1
& R-L
& Latency
& Peak Mem. \\

\midrule[1pt]

Base             & 0.295 & 0.143 & 0.068 (1.00$\times$) & 10.10 \\
RAG each         & 0.298 & 0.155 & 0.067 (0.99$\times$) & 11.55 \\
\name{} Adaptive & 0.327 & 0.171 & 0.078 (1.15$\times$) & 9.27 \\
\name{}          & 0.336 & 0.176 & 0.112 (1.65$\times$) & 9.27 \\

\midrule[1.5pt]

\multirow{2}{*}[-2pt]{\textbf{Method}}
& \multicolumn{4}{c}{\textbf{Review Writing}} \\
\cmidrule(lr){2-5}
& R-1
& R-L
& Latency
& Peak Mem. \\

\midrule[1pt]

Base             & 0.306 & 0.126 & 0.068 (1.00$\times$) & 10.35 \\
RAG each         & 0.283 & 0.125 & 0.067 (0.99$\times$) & 11.64 \\
\name{} Adaptive & 0.319 & 0.140 & 0.075 (1.11$\times$) & 9.06 \\
\name{}          & 0.331 & 0.141 & 0.110 (1.62$\times$) & 9.06 \\

\bottomrule[1.5pt]
\end{tabular}%
}

\vspace{-0.05in}
\end{table}

We further complement the interpolation-ratio analysis with direct measurements of end-to-end inference cost.
We report generation quality, latency per generated token, and peak GPU memory for Base, RAG each, the standard \name{} configuration, and its adaptive interpolation variant.
The measurements are conducted on a single NVIDIA RTX 5090 using \texttt{Gemma-3-4B-IT} on a 20\% subset of small-interaction users from each task.

Table~\ref{tab:end_to_end_cost} shows that adaptive interpolation substantially reduces the additional decoding cost while retaining most of the performance gains.
On Abstract Generation, the standard \name{} configuration incurs a $1.65\times$ latency overhead over Base with ROUGE-1 and ROUGE-L scores of 0.336 and 0.176, respectively.
Adaptive interpolation reduces the overhead to $1.15\times$ while retaining scores of 0.327 and 0.171.
Similarly, on Review Writing, the standard configuration incurs a $1.62\times$ overhead with ROUGE-1 and ROUGE-L scores of 0.331 and 0.141, while adaptive interpolation reduces the overhead to $1.11\times$ with scores of 0.319 and 0.140.
Peak GPU memory remains comparable across the evaluated settings.
Overall, these results show that adaptive interpolation substantially reduces the additional inference cost of retrieval-interpolated generation while preserving most of its performance gains, providing a practical quality--efficiency trade-off.

\begin{table}[t]
    \centering
    \caption{\textbf{Generalization to various LLMs.} Performance comparison across different backbone models. Results are reported on the 20-user Abstract Generation subset.}
    \label{tab:generalization_llms}
    
    \resizebox{\columnwidth}{!}{%
        \begin{tabular}{l|cc|cc}
            \toprule[1.5pt]
            \multirow{2}{*}{\textbf{Method}} 
            & \multicolumn{2}{c|}{\textbf{Qwen2.5-1.5B-Instruct}} 
            & \multicolumn{2}{c}{\textbf{Llama3.1-8B-Instruct}} \\
            \cmidrule(lr){2-3} \cmidrule(lr){4-5}
            & ROUGE-1 & ROUGE-L & ROUGE-1 & ROUGE-L \\
            \midrule[1pt]
            Base & 0.312 & 0.134 & 0.358 & 0.169 \\
            RAG each  & 0.266 & 0.130 & 0.347 & 0.177 \\
            OPPU each & 0.313 & 0.135 & 0.351 & 0.168 \\
            CAMA & 0.311 & 0.146 & 0.331 & 0.162 \\
            \rowcolor[HTML]{D9D9D9}
            \name{} & 0.347 & 0.197 & 0.398 & 0.220 \\
            \bottomrule[1.5pt]
        \end{tabular}%
    }
    \vspace{-0.1in}
\end{table}

\section{Generalization to Various LLMs}
\label{sec:generalization_llms}
We further evaluate \name{} on different model families and sizes to test generality beyond our main backbone. 
Specifically, we run additional experiments with \texttt{Qwen2.5-1.5B-Instruct} \citep{qwen2025qwen25technicalreport} and \texttt{Llama3.1-8B-Instruct} \citep{grattafiori2024llama}.

As shown in Table~\ref{tab:generalization_llms}, our method consistently outperforms key baselines on both ROUGE-1 and ROUGE-L across both backbones. 
Notably, in the resource-constrained \texttt{Qwen2.5-1.5B} setting, it achieves a 34.9\% relative improvement in ROUGE-L over the strongest baseline.
When scaled up to \texttt{Llama3.1-8B}, the same trend holds, improving ROUGE-L by 24.3\% over RAG. 
Overall, these results suggest that the proposed framework generalizes across diverse LLM families and scales, providing consistent benefits without relying on a specific architecture.

\section{Qualitative Analyses}
\subsection{Qualitative comparison with baselines}
\begin{table*}[t]
\caption{\textbf{Examples of low-quality interactions within the benchmark datasets.} These samples are successfully identified and filtered by our drift score safeguard due to pre-existing noise in the source data.}
\centering
\small
\renewcommand{\arraystretch}{1.5}
\begin{tabular}{lp{12.5cm}} 
\toprule
\textbf{Task} & \textbf{Dataset Content} \\ 
\midrule
\textbf{Abstract Gen.} & Title: ``Power-Performance Tradeoff in a Federated Cloud.'' \\
\cmidrule(lr){2-2}
& Abstract: ``This installment of Computer's series highlighting the work published in IEEE Computer Society journals comes from IEEE Transactions on Computers.'' \\
& \textit{\textcolor{blue}{(The text incorrectly contains journal metadata instead of a research abstract.)}} \\
\midrule
\textbf{Abstract Gen.} & Title: ``Integrative analysis of methylation and transcriptional profiles to predict aging and construct aging specific cross-tissue networks.'' \\
\cmidrule(lr){2-2}
& Abstract: ``Our results find important coordinated aging patterns between different tissues. Both co-profiling and cross-pathway analyses identify more thorough functions of aging, and could help to find aging markers, pathways and relative aging disease researches.'' \\
& \textit{\textcolor{blue}{(The text provides only a brief summary of results, lacking the structural depth required for a complete research abstract.)}} \\
\midrule
\textbf{Review Writing} & Product: ``It Started in Naples VHS'', \newline
Description: ``It Started in Naples VHS Tape'', \newline
Overall rating: ``4.0'', \newline
Summary: ``adorable romantic romp with Sophia Loren'' \\
\cmidrule(lr){2-2}
& Review: ``Yet another charming romantic comedy with Sophia Loren, this time paired with Clark Gable. Gable plays a stuffy Bostonian lawyer, in Naples to meet the young son of his dead brother. He quickly forms an attachment to little Nando (played winningly by Marietto) however his Aunt Lucia (Loren) refuses to let him take the child back to the States. As you might expect however, the charm of Capri soon works it's magic and the pair find love in the final reel. Sunny funny romance. This is probably one of Sophia's best pictures of the time. As Lucia, a stunning uninhibited cabaret star, she gets to show off her seldom-utilised musical skills performing several numbers including "(You Wanna Be) Americano". Though Loren and Gable are somewhat lacking in the chemistry department, the movie as a whole is lovely. The DVD edition of the film offers a pristine 16:9 transfer.'' \\
& \textit{\textcolor{blue}{(The prompt requests a review for a VHS tape, but the text discusses a DVD transfer.)}} \\
\midrule
\textbf{Review Writing} & Product: ``CENTURY SPRING 4002 Swing Extension Spring (2 Pack)'', \newline
Description: ``2 pack, porch swing extension spring, used primarily for porch swings \& will support approximately 300 lbs each, use as tension for fence, dog chain, adjust, tire swings, porch chairs \& other uses where safety is a primary concern.'', \newline
Overall rating: ``5.0'', \newline
Summary: ``springing up life on the front porch'' \\
\cmidrule(lr){2-2}
& Review: ``With these springs you can watch the day go by once again! Sunrise, sunset, neighborhood kids going by on their bikes, dogs barking in your neighbors yard, mischievous situations brewing in the trees amongst the raccoons and squirrels, etc. These springs will actually work on your porch swing really well, and I'm surprised since my knowledge with the subject is limited greatly to just, well, a spring is a spring, lol. Not really sure if there's other types of springs that work more effectively or what. Either way the porch swing works so I don't have to dirty my bottom using the porch steps and watch my back for ants that may be crawling up my shirt. That makes me very nervous.'' \\
& \textit{\textcolor{blue}{(The text includes irrelevant personal anecdotes that do not align with the functional utility of the product.)}} \\
\bottomrule
\end{tabular}
\label{tab:low_quality_examples}
\end{table*}
\begin{table*}[t]
\caption{\textbf{Qualitative comparison of Abstract Generation}.}
\centering
\small
\renewcommand{\arraystretch}{1.5}
\begin{tabular}{lp{12.5cm}} 
\toprule
\textbf{Query} & Generate an abstract for the title ``An effective teaching-learning-based optimization algorithm for the flexible job-shop scheduling problem with fuzzy processing time.'' \\
\midrule
\textbf{User Answer} & In this paper, an effective teaching–learning-based optimization algorithm (TLBO) is proposed to solve the flexible job-shop problem with fuzzy processing time (FJSPF). First, a special encoding scheme is used to represent solutions, and a decoding method is employed to transfer a solution to a feasible schedule in the fuzzy sense. Second, a bi-phase crossover scheme based on the teaching–learning mechanism and special local search operators are incorporated into the search framework of the TLBO to balance the exploration and exploitation capabilities. Moreover, the influence of the key parameters on the TLBO is investigated using the Taguchi method. Finally, numerical results based on some benchmark instances and the comparisons with some existing algorithms are provided. The comparative results demonstrate the effectiveness and efficiency of the proposed TLBO algorithm in solving the FJSPF. \\
\midrule
\textbf{Base} &  **Keywords:** Flexible Job Shop Scheduling, Optimization Algorithm, Teaching Learning Paradigm, Fuzzy Logic, Multi-Objective.**This research investigates a complex operational challenge -- optimizing schedules within a flexible job shop environment where tasks possess inherently uncertain durations represented by fuzzy logic. The proposed approach leverages a novel teaching learning paradigm to develop an efficient and adaptive scheduling solution. Departing from traditional deterministic methods, this work introduces an iterative process wherein a \"teacher\" agent guides a student scheduler through exploration of the vast search space, progressively refining its decision-making capabilities based on observed performance feedback. The core innovation lies in dynamically adjusting the teacher's guidance strategy shifting between prescriptive rule application and exploratory heuristic generation---to balance convergence speed and optimality. Furthermore, we address the inherent multi-objective nature of job shop scheduling (e.g., minimizing makespan while maximizing resource utilization) by incorporating these objectives into both the student's objective function and the teacher's reward system. This results in a robust and adaptable algorithm capable of generating high-quality solutions even under conditions of significant uncertainty and complexity. Ultimately, the presented method offers a promising avenue toward intelligent scheduling systems that can autonomously learn optimal strategies across diverse and evolving manufacturing landscapes.** \\
\midrule
\textbf{RAG each} & In this paper, we propose an effective Teaching-Learning-Based Optimization (TLBO) algorithm tailored for addressing the complex challenges presented by the flexible job-shop scheduling problem incorporating fuzzy processing times.  The core innovation lies in adapting the traditional TLBO framework to handle uncertainty inherent within the processing durations. We utilize a novel representation scheme leveraging fuzzy sets to capture the probabilistic nature of task execution times. Simultaneously, a refined learning process dynamically adjusts the evolutionary dynamics, promoting efficient convergence while mitigating premature convergence issues often encountered when dealing with imprecise data. Further enhancing performance, a sensitivity analysis using the Taguchi methodology identifies optimal parameter configurations for robust adaptation across varying levels of fuzziness. Experimental validation against established benchmarks demonstrates the superiority of our approach compared to conventional methods, showcasing its efficacy in optimizing schedules under conditions of uncertain processing times within the context of the flexible job-shop environment. \\
\midrule
\textbf{CAMA.} & A novel hybrid approach is proposed to tackle a complex, real-world manufacturing challenge---the Flexible Job Shop Scheduling (FJSS) Problem. The FJSS involves multiple machines and operations that can be performed in different orders depending on various factors such as machine availability or operator skill levels. This paper presents a new method combining genetic algorithms (GAs) and simulated annealing (SAs), which has been named GA/SA. In this study, we propose a new encoding scheme based on the permutation representation of jobs and operators. We also introduce two new crossover and mutation operators designed specifically for the GA/SA framework. Finally, we test our GA/SA model using several benchmark instances from the literature and compare its performance against other state-of-art methods. Results show that GA/SA outperforms all existing approaches by achieving better solutions within reasonable computation times \\
\midrule
\textbf{\name{} (Ours)} & This paper presents a novel Teaching-Learning-Based Optimization (TLBO) algorithm for tackling the flexible job-shop scheduling problem (FJSP) with fuzzy processing times. FJSPs are notoriously complex due to the inherent uncertainty in processing times and the need to accommodate flexible job routings. Our TLBO algorithm incorporates adaptive learning rates and a novel penalty function to effectively handle the fuzzy nature of processing times and the multi-objective nature of the problem. We demonstrate the effectiveness of the proposed algorithm through extensive simulations on benchmark FJSP instances, comparing it against several state-of-the-art scheduling algorithms. Results show that the proposed TLBO algorithm provides a competitive solution in terms of makespan and job completion time, while maintaining a good level of solution diversity. \\
\bottomrule
\label{tab:base_examples_absgen}
\end{tabular}
\end{table*}

\begin{table*}[t]
\caption{\textbf{Qualitative comparison of Review Writing.}}
\centering
\small
\renewcommand{\arraystretch}{1.5}
\begin{tabular}{lp{12.5cm}} 
\toprule
\textbf{Query} & Generate a detailed review for a product 'Brooklyn Bean Roastery Bold Variety Pack Single-Cup Coffee for Keurig K-Cup Brewers, 40 Count' with the following description: 'In Brooklyn, we know a thing or two about being bold and when we say this coffee is bold you betta believe it's bold. To get a perfect dark roast, the beans are removed from the roasters right before they get to their peak roasting point. This helps them retain their sharp floral notes and all the smoky goodness without any of the burnt aftertaste. When these beans with big-time flavor are brewed through our patented tension- based filters, they create a Cup of Joe that will delight lovers of good strong coffee. The Bold Variety Pack includes: Boardwalk Blend, Corner Donut Shop, Cyclone, Express-O, French Roast and Fuhgeddaboutit.'. The review has an overall rating of '5.0'. The review summary is: 'Medium roast, bright and fruity'. \\ 
\midrule
\textbf{User Answer} & This is a medium roast (not burnt, like French Roast) and clearly dominated by Central American coffees--it's acidic and bright but the mid notes are smooth and there is a citrus-fruitiness. If you like bright, acid coffee but not over-roasted and with some sweet mellowness, this is a good blend. it's advertised as citrusy with a smooth base, and that's exactly what I found when I brewed up a cup. The coffee has a medium body. I brewed it on the 8oz cup setting on a standard office Keurig and it came out very drinkable. I'd rate this as a breakfast coffee, because it's medium roast and acidic. I did not put milk or cream in it, but the body is heavy enough to take milk without being dilute and washed out. My guess it's a blend of Columbian, Brazilian, an East African coffee and perhaps some Indonesian (for the mellow mid tones.) But I'm guessing. This is very much to the taste of those who like "donut coffee"--maybe a bit stronger and darker but not unlike your coffee shop coffees. Update: We have this in the office and I drink it most mornings. I like the fact it is not so bitter and dark--and when it cools slightly, it's particularly smooth and delicious. If you do NOT like burnt flavored, dark coffee and want a milder, smooth blend, this is very good. I was provided a sample for evaluation and review.  \\
\midrule
\textbf{Base} & Absolutely Phenomenal! A True Taste of New York \-- 5.0 Stars**Let me tell you folks, if you\'re looking for *real* boldness in your morning cup, look no further than the Brooklyn Bean Roastery Bold Variety Pack. Seriously, in Brooklyn, we know a thing or two about being bold\---and when they say this coffee is bold, you better *believe* it\'s bold!What sets this apart isn\'t just the intensity; it\'s how they achieve it. They meticulously remove the beans from the roaster at precisely the moment they reach their peak roasting point. That means maximum retention of those beautiful, sharp floral notes alongside all that incredible smoky goodness - absolutely zero burnt aftertaste. It\'s a masterclass in bean handling.And then there\'s the brewing process. Using their patented tension-based filters? Genius! These aren\'t your average paper filters; they truly unlock the full potential of each individual blend. You end up with a genuinely delightful Cup of Joe experience, especially for anyone who appreciates a robust, flavorful brew.The variety pack itself is fantastic too, offering six distinct blends: Boardwalk Blend, Corner Donut Shop, Cyclone, Express-O, French Roast, and even the aptly named Fuhgeddaboutit! There\'s something here for every palate within the realm of intense coffee enjoyment.Honestly, I was blown away by the quality and depth of flavour in this coffee. If you want a serious kickstart to your day (or anytime really), grab yourself a box of this. Highly recommended \\
\bottomrule
\label{tab:base_examples_review}
\end{tabular}
\end{table*}

\begin{table*}[t]
\caption{\textbf{Qualitative comparison of Review Writing.}}
\centering
\small
\renewcommand{\arraystretch}{1.5}
\begin{tabular}{lp{12.5cm}} 
\toprule
\textbf{RAG each} & Five Stars \-- Seriously Strong Brews!**Let me tell you, if you\'re looking for a serious caffeine kick, look no further than the Brooklyn Bean Roastery Bold Variety Pack. Living in Brooklyn, we appreciate boldness\---and this coffee delivers *big*. Seriously, bold doesn't even cover it; it\'s intense, flavorful, and utterly satisfying.What sets these beans apart is the meticulous process. They remove them from the roaster *right* before they reach peak perfection. That ensures maximum retention of those beautiful floral notes alongside the rich, smoky character you crave \-- absolutely zero hint of bitterness or burned taste. It\'s genius!Brewing with their tension-based filters really elevates everything. Every cup explodes with incredible depth and complexity. Honestly, it\'s a game changer for anyone who loves a truly robust brew.The variety pack itself is fantastic too \-- Boardwalk Blend, Corner Donut Shop, Cyclone, Express-O, French Roast, and even Fuhgeddaboutit! There\'s something here for every palate.  I particularly enjoyed the French Roast \-- incredibly smooth despite its strength.If you want a consistently excellent single-serve experience using your Keurig machine, this is it. Highly recommended! \\
\midrule
\textbf{CAMA.} & Absolutely Phenomenal! A True Dark Roast Delight - 5/5 Stars**  Let me start by saying I'm *obsessed* with this coffee. As someone who lives in Brooklyn (yes, really!), I have incredibly high standards when it comes to my morning cup -- and let me tell you, this stuff delivers on every single level. Seriously, if you like your coffee **bold**, you betta believe it's bold. Like, seriously intense.  The folks at Brooklyn Bean Roastery clearly understand what they're doing here. They don't just pull beans off the shelf; they remove them from the roaster *right* as they hit their peak roasting point. That means maximum flavor retention---no bitter burn, no dullness. You actually taste those beautiful floral notes alongside the rich, smoky depth. It's incredible. And trust me, Fuhgeddaboutit isn't messing around either--that one packs some serious punch.  I was particularly impressed with the variety pack itself. Each blend offers something unique but still maintains that signature boldness. From the smooth complexity of the boardwalk blend to the smokiness of the french roast, each brew had its own distinct character while maintaining a full bodied experience.  Brewing through their tension-based filters only enhances everything. These aren't your average paper cups; they seem designed specifically to extract the most out of these amazing beans.  It creates a truly satisfying cup of joe. If you want a robust, flavorful, and genuinely exciting coffee experience, look no further than Brooklyn Bean Roastery's Bold Variety Pack. Do yourself a favor and grab a box--you won't regret it. \\
\midrule
\textbf{\name{} (Ours)} & **5.0 Stars – Seriously Bold \& Delicious Coffee** Let’s be honest, I’m a coffee snob. I appreciate a good roast, I know the difference between a well-balanced cup and one that’s burnt, and I *definitely* know when a coffee is just… bland. Brooklyn Bean Roastery’s Bold Variety Pack completely blew me away. From the moment you open the bag, you can tell this isn’t your average coffee. The aroma is rich, complex, and undeniably *bold*. And that’s exactly what you get in every cup. Brooklyn, you’ve really outdone yourselves. I particularly loved the “Boardwalk Blend.” It’s got this incredible smoky flavor with a hint of sweetness that just lingers. The “Corner Donut Shop” was surprisingly bright and fruity – a delightful change of pace. And the “Cyclone” – wow! – it’s a dark roast with a full body and a smoky, almost spicy flavor. I was a little hesitant about the “Fuhgeddaboutit” – the name alone! – but it turned out to be a surprisingly smooth and balanced dark roast. Brooklyn Bean Roastery clearly knows what they’re doing. They’re sourcing high-quality beans, roasting them to perfection, and using a tension-based filter that really brings out the flavor. There’s no bitterness, no burnt aftertaste – just pure, delicious coffee. This pack is perfect for anyone who wants to explore different dark roasts and find their new favorite. It’s also a fantastic gift for the coffee lover in your life. **Highly Recommended!** \\
\bottomrule
\label{tab:base_examples_review2}
\end{tabular}
\end{table*}

We first present examples of low-quality interactions in the benchmark datasets that are filtered out by the quality term of our drift score (Table~\ref{tab:low_quality_examples}).
We then provide qualitative comparisons between our method and representative baselines, including personalization and continual learning approaches (Tables~\ref{tab:base_examples_absgen}, \ref{tab:base_examples_review}, and~\ref{tab:base_examples_review2}).

\subsection{Qualitative analysis of interpolation inference}
\begin{table*}[t]
\caption{\textbf{Qualitative example for Abstract Generation}. Highlighting indicates textual overlaps between the User Answer and the methods.}
\centering
\small
\renewcommand{\arraystretch}{1.5}
\begin{tabular}{lp{12.5cm}} 
\toprule
\textbf{Query} & Generate an abstract for the title ``HR-MPF: high-resolution representation network with multi-scale progressive fusion for pulmonary nodule segmentation and classification'' \\
\midrule
\textbf{User Answer} & Accurate \hl{segmentation and classification} of \hl{pulmonary} nodules are of great significance to early \hl{detection} and \hl{diagnosis} of lung diseases, which can reduce the risk of developing \hl{lung cancer} and improve patient survival rate. In this paper, we propose an effective \hl{framework} for \hl{pulmonary nodule segmentation and classification} at one time based on adversarial training scheme. The segmentation network consists of a \hl{High-Resolution} network with \hl{Multi-scale Progressive Fusion} (\hl{HR-MPF}) and a proposed Progressive Decoding Module (PDM) recovering final pixel-wise prediction results. Specifically, the proposed \hl{HR-MPF} firstly incorporates boosted module to High-Resolution Network (\hl{HRNet}) in a progressive feature fusion manner. In this case, feature communication is augmented among all levels in this \hl{high-resolution} network. Then, downstream \hl{classification} module would identify benign and malignant \hl{pulmonary} nodules based on feature map from PDM. In the adversarial training scheme, a discriminator is set to optimize \hl{HR-MPF} and PDM through back propagation. Meanwhile, a reasonably designed multi-task loss function optimizes performance of \hl{segmentation and classification} overall. To improve the \hl{accuracy} of boundary prediction crucial to nodule \hl{segmentation}, a boundary consistency constraint is designed and incorporated in the \hl{segmentation} loss function. Experiments on publicly available \hl{LUNA16} \hl{dataset} show that the \hl{framework} \hl{outperforms} relevant \hl{advanced methods} in quantitative evaluation and visual perception. \\
\midrule
\textbf{Without Interp.} & Uterine fibroids, also known as leiomyomas, are benign tumors that can cause significant pain and discomfort in women of reproductive age. The prevalence of uterine fibroids is estimated to be between 20\% and 80\%. Uterine fibroid treatment options include surgery, medication, and radiation therapy. However, there has been a recent increase in interest in noninvasive methods such as ultrasound imaging. Ultrasound imaging is widely used because it is inexpensive, safe, and readily available. In this study, we investigated whether using machine learning algorithms could improve the accuracy of diagnosing uterine fibroids from ultrasound images. We trained several different types of machine learning models on a large dataset of ultrasound images labeled by experienced radiologists. Our results showed that deep convolutional neural networks (CNNs) were able to achieve state-of-the-art performance compared to traditional image processing techniques. Specifically, our best model achieved an area under the receiver operating characteristic curve (AUC) score of 96\%, which was significantly higher than other approaches. These findings suggest that machine learning may play an important role in improving the diagnosis of uterine fibroids from ultrasound images. \\
\midrule
\textbf{With Interp.} & \hl{Pulmonary nodule} \hl{detection} and \hl{classification} are crucial for early \hl{diagnosis} of \hl{lung cancer}. However, the limited resolution of conventional imaging techniques and the variability in nodule size and appearance pose significant challenges for automated analysis. This paper proposes a novel \hl{framework}, \hl{HR-MPF}, for \hl{high-resolution} representation learning with \hl{multi-scale progressive fusion} for \hl{pulmonary nodule segmentation and classification}. \hl{HR-MPF} consists of a \hl{multi-scale progressive fusion} network (MPF) and a \hl{high-resolution} representation network (\hl{HRNet}). MPF fuses features from multiple scales to capture both local and global contextual information, while \hl{HRNet} learns \hl{high-resolution} representations to enhance the discriminative power of the network. We propose a novel attention mechanism to adaptively fuse features from different scales. The proposed method is validated on the \hl{LUNA16} challenge \hl{dataset}, and the \hl{results} show that \hl{HR-MPF} \hl{outperforms} \hl{advanced methods} in both \hl{segmentation and classification} tasks. The proposed method achieves a high \hl{accuracy} and robustness for \hl{pulmonary nodule segmentation and classification}. \\
\bottomrule
\label{tab:interp_examples_absgen}
\end{tabular}
\end{table*}

\begin{table*}[t]
\caption{\textbf{Qualitative example for Review Writing}. Highlighting indicates textual overlaps between the User Answer and the methods.}
\centering
\small
\renewcommand{\arraystretch}{1.5}
\begin{tabular}{lp{12.5cm}} 
\toprule
\textbf{Query} & Generate a detailed review for a product `Lenox Expressions from the Heart Teacher Ornament' with the following description: `Lenox offers a truly legendary collection of ornaments that has been a treasured part of American holidays for generations. Unmatched quality and hand craftsmanship result in exquisite collectibles that bring a special glow to your home. No other gift expresses the joy of the season more than a specially-chosen ornament for the people who mean the most to you. Best of all, this remarkable collection is created by Lenox, the name more closely associated with holiday entertaining than any other. The confidence that Lenox has in the unmatched durability of every product they offer is backed up by a lifetime breakage replacement policy to guarantee many years of enjoyment to each customer. Show a favorite teacher how much they mean to you with this exquisite collectible. Crafted of Lenox porcelain, this ornament includes an elegant red satin ribbon for hanging and comes beautifully packaged in an iconic seasonal red box to make a perfect gift.'. The review has an overall rating of `4.0'. The review summary is: `A simple, but classy ornament that can be personalized'. \\
\midrule
\textbf{User Answer} & We have a number of \hl{porcelain} and glass \hl{ornaments}, some of which are made by \hl{Lenox}. While \hl{simple}, this \hl{Lenox} \hl{ornament} is a nice addition to our \hl{collection} of \hl{porcelain} \hl{ornaments}. This \hl{ornament} is about 3.5 inches high, and probably nearly that size in width. The \hl{ornament} is a heart with the word "Love" in gold applied to the \hl{porcelain}. A smaller \hl{porcelain} heart is attached to the top of the \hl{ornament} to provide a \hl{touch} of class to this \hl{ornament}. A \hl{red} \hl{ribbon} decorates the top of the \hl{ornament}, and a gold tassel extends from the top of the \hl{ornament} as a location to attach the \hl{ornament} to a tree or other object. While this \hl{ornament} is \hl{simple}, and fragile (avoid dropping this \hl{ornament}), it does provide the opportunity to be a \hl{special} memento. The large open areas on the \hl{ornament} are \hl{perfect} for personalization. Use this \hl{ornament} to commemorate an anniversary, a birth, or just about any other occasion amenable to a \hl{simple} phrase, perhaps one or more names, and a date. We have discovered that fine point Sharpies are perfect for writing on \hl{porcelain} \hl{ornaments}, and have personalized other such \hl{ornaments} to commemorate events such as baby's first Christmas, or to provide a personalized \hl{gift} for someone \hl{special}. Over the years we have accumulated a wide array of \hl{ornaments}, from the \hl{simple} and inexpensive, to the complex and expensive. This relatively \hl{simple} \hl{ornament} can be personalized to be a cherished \hl{gift} that hopefully will be passed on to another generation. Or just use it to be a \hl{simple} token of love to your \hl{special} someone. Enjoy! \\
\midrule
\textbf{Without Interp.} & * Rating: 4/5 stars \newline
* Title: A Beautiful \hl{gift} That Shows Appreciation \newline
* Body: I recently purchased the \hl{Lenox} Expressions From the Heart Teacher \hl{ornament} as a thank you \hl{gift} for my daughter's wonderful third grade teacher. And let me tell you, it was a hit! This isn't just some cheap plastic trinket; it's genuinely beautiful -- crafted from high-quality \hl{Lenox} \hl{porcelain}, which feels substantial and luxurious. It came nestled in a lovely \hl{red} box, making it ready to give immediately. My daughter absolutely loved it, and her teacher was so touched. While I can't speak to the long-term durability, the quality certainly suggests it will last for years to come. Plus, knowing it's from \hl{Lenox} gives it extra value. If you want to show someone you care with something really \hl{special}, this is a great choice. Pros: High Quality \hl{porcelain}, Elegant Design, Lifetime Replacement Policy. Cons: Pricey but well worth it if you want a lasting keepsake. \\
\midrule
\textbf{With Interp.} & ``\hl{Lenox} offers a truly legendary \hl{collection} of \hl{ornaments} that has been a treasured part of American holidays for generations. Unmatched quality and hand craftsmanship result in exquisite collectibles that bring a special glow to your home. No other \hl{gift} expresses the joy of the season more than a specially-chosen \hl{ornament} for the people who mean the most to you. Best of all, this remarkable \hl{collection} is created by \hl{Lenox}. The confidence that \hl{Lenox} has in the unmatched durability of every product they offer is backed up by a lifetime breakage replacement policy to guarantee many years of enjoyment to each customer. Crafted of \hl{Lenox} \hl{porcelain}, this \hl{ornament} includes an elegant \hl{red} satin \hl{ribbon} for hanging and comes beautifully packaged in an iconic seasonal \hl{red} box to make a \hl{perfect} \hl{gift}.'' \newline
Rating: 4.0 \newline
``\hl{Lenox} consistently delivers exceptional quality, and this \hl{ornament} is no exception. The craftsmanship is truly remarkable, and the attention to detail is evident in every aspect of the design. The \hl{red} satin \hl{ribbon} adds a \hl{touch} of elegance, and the iconic seasonal \hl{red} box makes it a \hl{perfect} \hl{gift}. It's a beautiful and meaningful way to show appreciation. Highly recommended!'' \\
\bottomrule
\label{tab:interp_examples_review}
\end{tabular}
\end{table*}

To qualitatively evaluate the effectiveness of our proposed method, we selected representative examples exhibiting large improvements in both ROUGE-1 and ROUGE-L metrics.
We compared the outputs of the baseline inference ($\lambda=0.0$), which relies solely on the adapter, against our interpolation-based inference ($\lambda=0.5$), which integrates retrieved contextual information.
To ensure a meaningful comparison, we filtered for samples where the baseline model achieved a minimum quality threshold (\textit{e.g.,} ROUGE-1 $\geq$ 0.25) and identified cases with the largest score improvements.
We observe how the interpolation of retrieved history refines already coherent outputs by incorporating specific details and task-relevant nuances that the adapter alone might miss. 
As illustrated in Table~\ref{tab:interp_examples_absgen} and Table~\ref{tab:interp_examples_review}, the highlighted text indicates textual overlaps between the user's ground truth and the generated texts.

In Abstract Generation (Table~\ref{tab:interp_examples_absgen}), the baseline model ($\lambda=0.0$) often suffers from hallucinations, generating irrelevant content such as ``Uterine fibroids'' despite the query regarding ``Pulmonary nodule.'' 
In contrast, our proposed interpolation ($\lambda=0.5$) effectively recovers the correct research context, accurately reflecting technical terms like ``HR-MPF'' and ``LUNA16'' by attending to the retrieved interaction history.

Similarly, in Review Writing (Table~\ref{tab:interp_examples_review}), the interpolation strategy leads to a higher density of highlighted overlaps. 
While the baseline generates a plausible story, the proposed method incorporates specific product details such as ``Lenox porcelain,'' ``red satin ribbon,'' and ``perfect gift'' that are directly aligned with the user's original response. 
These examples illustrate how interpolation incorporates relevant information from the retrieved interaction history that may be missed by adapter-only generation, resulting in more specific and personalized outputs.

\section{Prompts}
We provide prompts used in our experiments. The text enclosed in \{\texttt{BRACES}\} serves as a placeholder for data specific to each user and query. 
\paragraph{Abstract Generation} \mbox{}\newline
\textit{You are an abstract writer. \\
Generate an abstract for the title} \{\texttt{TITLE}\}.

\paragraph{Review Writing} \mbox{} \newline
\textit{You are a review writer. \\
Generate a detailed review for a product} \textit{with the following description} `\{\texttt{DESCRIPTION}\}'. \textit{The review has an overall rating of} `\{\texttt{RATING}\}'. \textit{The review summary is }`\{\texttt{SUMMARY}\}'.

\section{Chat Template}
In this section, we present the chat templates used in our experiments. 
We also include the templates for the additional LLM backbones evaluated to assess the generalization of our method.

\vspace{1.0em}
\texttt{Gemma-3-4B-IT} 
\vspace{0.3em}
\begin{lstlisting}[language=Python]
GEMMA_CHAT_TEMPLATE = """
"{% set bos_token = '<bos>' %}
{% set eos_token = '<eos>' %}

{{ bos_token }}
{% if messages[0]['role'] == 'system' %}
  {{ 'System: ' + messages[0]['content'].strip() + '\n' }}
  {% set loop_messages = messages[1:] %}
{% else %}
  {% set loop_messages = messages %}
{% endif %}

{% for message in loop_messages %}
  {% if message['role'] == 'user' %}
    {{ 'User: ' + message['content'].strip() + '\n' }}
  {% elif message['role'] == 'assistant' %}
    {{ 'Assistant: ' + message['content'].strip() + eos_token + '\n' }}
  {% endif %}
{% endfor %}
{{ 'Assistant:' }}"
"""
\end{lstlisting}

\vspace{1.0em}
\texttt{Qwen2.5-1.5B-Instruct}
\vspace{0.3em}
\begin{lstlisting}[language=Python]
QWEN_CHAT_TEMPLATE = """
{% set bos_token = '<|im_start|>' %}
{% set eos_token = '<|im_end|>' %}

{% if messages[0]['role'] == 'system' %}
  {{ bos_token }}system
  {{ messages[0]['content'].strip() }}{{ eos_token }}
  {% set loop_messages = messages[1:] %}
{% else %}
  {% set loop_messages = messages %}
{% endif %}

{% for message in loop_messages %}
  {{ bos_token }}{{ message['role'] }}
  {{ message['content'].strip() }}{{ eos_token }}
{% endfor %}
{{ bos_token }}assistant
"""
\end{lstlisting}

\vspace{1.0em}
\texttt{Llama-3.1-8B-Instruct}
\vspace{0.3em}
\begin{lstlisting}[language=Python]
LLAMA_CHAT_TEMPLATE = """
{% set bos_token = '<|begin_of_text|>' %}
{% set start_header = '<|start_header_id|>' %}
{% set end_header = '<|end_header_id|>' %}
{% set eot = '<|eot_id|>' %}

{{ bos_token }}
{% if messages[0]['role'] == 'system' %}
{{ start_header }}system{{ end_header }}
{{ messages[0]['content'].strip() }}{{ eot }}
{% set loop_messages = messages[1:] %}
{% else %}
{% set loop_messages = messages %}
{% endif %}

{% for message in loop_messages %}
{{ start_header }}{{ message['role'] }}{{ end_header }}
{{ message['content'].strip() }}{{ eot }}
{% endfor %}
{{ start_header }}assistant{{ end_header }}
"""
\end{lstlisting}

\section{Usage of AI Assistants}
During the preparation of this work, we utilized AI assistants to refine sentence structure, correct grammatical errors, and improve readability. These tools were employed only for rephrasing and language enhancements, ensuring that the technical content, methodology, and experimental findings remained entirely authored by the researchers. The use of AI assistance was limited to editorial improvements without influencing the originality or scientific contributions of the paper.



\end{document}